\documentclass{article}

\usepackage{microtype}
\usepackage{graphicx}
\usepackage{subfigure}
\usepackage{booktabs} 
\usepackage{float}
\usepackage{tikz}
\usepackage{setspace}
\usepackage{bm}
\usepackage{amssymb}
\usepackage{amsmath}
\usepackage{multicol}
\usepackage{graphicx}
\usepackage{array}
\usepackage{mathrsfs}
\usepackage{appendix}
\usepackage{booktabs}
\usepackage{url}
\usepackage{comment}
\usepackage{hhline}
\newtheorem{theorem}{Theorem}
\newtheorem{definition}[theorem]{Definition}
\newtheorem{lemma}{Lemma}

\usepackage{todonotes}
\usepackage[T1]{fontenc}
\usepackage[utf8]{inputenc}
\usepackage{authblk}

\usepackage{hyperref}

\usepackage[accepted]{icml2019}

\icmltitlerunning{On Negative Transfer and Structure of Latent Functions}

\begin{document}

\twocolumn[
\icmltitle{On Negative Transfer and Structure of Latent Functions \\ in Multi-output Gaussian Processes}



\icmlsetsymbol{equal}{*}
\begin{icmlauthorlist}
\icmlauthor{Moyan Li}{to}
\icmlauthor{Raed Kontar}{to}
\end{icmlauthorlist}

\icmlaffiliation{to}{Department of Industrial \& Operations Engineering, University of Michigan, Ann Arbor, MI}

\icmlcorrespondingauthor{Raed Kontar}{alkontar@umich.edu}
\icmlkeywords{Machine Learning, ICML}

\vskip 0.3in]


\printAffiliationsAndNotice{}  
\begin{abstract}
The multi-output Gaussian process ($\mathcal{MGP}$) is based on the assumption that outputs share commonalities, however, if this assumption does not hold negative transfer will lead to decreased performance relative to learning outputs independently or in subsets. In this article, we first define negative transfer in the context of an $\mathcal{MGP}$ and then derive necessary conditions for an $\mathcal{MGP}$ model to avoid negative transfer. Specifically, under the convolution construction, we show that avoiding negative transfer is mainly dependent on having a sufficient number of latent functions $Q$ regardless of the flexibility of the kernel or inference procedure used. However, a slight increase in $Q$ leads to a large increase in the number of parameters to be estimated. To this end, we propose two latent structures that scale to arbitrarily large datasets, can avoid negative transfer and allow any kernel or sparse approximations to be used within. These structures also allow regularization which can provide consistent and automatic selection of related outputs. 
\end{abstract}
\section{Introduction}
\label{sec:intro}
The multi-output, also referred to as multivariate/vector-valued/multitask, Gaussian process ($\mathcal{GP}$) draws it root from transfer learning, specifically multitask learning. The goal is to integratively analyze multiple outputs in order to leverage their commonalities and hence improve predictive and learning accuracy. Indeed, the multi-output $\mathcal{GP}$ ($\mathcal{MGP}$) has seen many success stories in the last decade. This success can be largely attributed to the convolution construction which provided the capability to account for heterogeneity and non-trivial commonalities in the outputs.

The convolution process ($\mathcal{CP}$) is based on the idea that a $\mathcal{GP}$, $f_i (\bm{x}): \mathcal{R}^\mathcal{D} \rightarrow \mathcal{R}$ can be constructed by convolving a latent Gaussian process $X(\bm{x})$ with a smoothing kernel $K_i(\bm{x})$. This construction, first proposed by \citet{ver1998constructing} and \citet{higdon2002space},  is equivalent to stimulating a linear filter characterized by the impulse response $K_i(\bm{x})$. The only restriction is that the filter is stable, i.e., $\int |K_i(\bm{u})|d\bm{u} < \infty$.  Given the $\mathcal{CP}$ construction, if we share multiple latent functions $X_q(\bm{x})$ across the outputs $f_i(\bm{x}), i\in \mathcal{I}$, then all outpouts can be expressed as jointly distributed $\mathcal{GP}$, i.e., an $\mathcal{MGP}$ \citep{alvarez2011computationally}. This is shown in (\ref{eq:cp}). \begin{equation}
\label{eq:cp}
f_i(\bm{x})=\sum_{q=1}^{Q} K_{qi}(\bm{x}) \star X_q(\bm{x}) ,
\end{equation}
where $\star$ denotes a convolution. The key feature in  (\ref{eq:cp}) is that it allows information to be shared through different kernels (with different parameters) which enables great flexibility in describing the data. {Further, many models used to build cross correlations across outputs including the large class of separable covariances and the linear model of corregonialization are special cases of the convolution construction \citep{alvarez2012kernels,fricker2013multivariate}}.

Since then, work on $\mathcal{MGP}$ has mainly focused on two trends: (1) \textit{Efficient inference procedures} that address the computational complexity (a challenge inherited from the $\mathcal{GP}$) \citep{wang2019exact,nychka2015multiresolution,gramacy2015local,damianou2016variational} . This literature has mainly focused on variational inference which laid the theoretical foundation for the commonly used class of inducing point/kernel approximation \citep{burt2019rates}. Interestingly, variational inference also reduced overfitting and helped generalization due to its regularizing impact. Some work in this area include: \citet{zhao2016variational,nguyen2014collaborative,moreno2018heterogeneous}, to name a few. (2) \textit{Building expressive kernels} \citep{parra2017spectral,chen2019multioutput,ulrich2015gp} that often can represent certain unique features of the data studied. Most of which are based on spectral kernels, despite the fact that convolved covariance based on the exponential, Gaussian or Mat\'ern kernels are still most common in applications. Relating to this, recent work has also studied a non-linear version of the convolution operator \citep{alvarez2018non}. 

However, a key question is yet to be answered. The $\mathcal{MGP}$ is based on the assumption that outputs share commonalities, but what happens if this assumption does not hold? would \textbf{negative transfer} occur? which in turn leads to \textit{forced correlation and decreased performance relative to learning each output independently} \citep{caruana1997multitask}. This question is especially relevant when using the $\mathcal{CP}$, which implicitly implies that outputs have heterogeneous, possibly unique, features. For instance, following recent literature, would an expressive kernel and an efficient inference procedure for finding good kernel parameter estimates automatically avoid spurious correlations? or say in an extreme case where all outputs share no commonalities, would the $\mathcal{MGP}$ automatically collapse into independent $\mathcal{GP}$s?. 

In this article we shed light on the aforementioned questions. Specifically, we first define negative transfer in the context of an $\mathcal{MGP}$. We then show that addressing the challenge above is mainly dependent on having a sufficient number of latent functions, i.e., $Q$ in (\ref{eq:cp}). However, even when $N$ is relatively small, a small increase in $Q$ would cause the number of parameters to be estimated to skyrocket. This renders estimation in such a non-convex and highly nonlinear setting impractical, which explains why current literature including the above  cited papers only use $1 \leq Q \leq 4$. To this end, we propose \textit{easy-to-implement relaxation models} on the $\mathcal{MGP}$ structure that scale to arbitrarily large datasets, can avoid negative transfer and allow any kernel or sparse approximations to be used within. A key feature of our models, is that they allow regularization penalties on the hyper-parameters which can provide consistent selection of related/unrelated outputs.  

We organize the remaining paper as follows. Sec. \ref{sec:prelim} provides some preliminaries. Sec. \ref{sec:define} defines negative transfer and provide necessary conditions to avoid it in the $\mathcal{MGP}$. In Sec. \ref{sec:relaxation} we provide scalable relaxation models on the $\mathcal{MGP}$ structure, we then explore regularization schemes that help generalization and automatic selection in the relaxation models. In Sec. \ref{sec:penalty} we provide a statistical guarantee on the performance of the penalized model.  A proof of concept and illustration on real-data is given in Sec. \ref{sec:case}. Finally, we conclude our paper in Sec. \ref{sec:conclusion}. Technical details and a detailed code are given in the supplementary materials.

\section{Preliminaries} \label{sec:prelim}

Consider the set of $N$ noisy output functions $\bm y (\bm x) = [y_1(\bm x),\cdots,y_N(\bm x)]^{\top}$ and let $\mathcal{I}=\{1,\cdots, N\}$ be the corresponding index set. 
\begin{equation*}
\label{eq:structure}
\begin{gathered}
\begin{bmatrix}
y_1(\bm{x})\\y_2(\bm{x})\\ \vdots \\y_N (\bm{x})
\end{bmatrix}
=
\begin{bmatrix}
f_1 (\bm{x})\\f_2(\bm{x})\\ \vdots \\f_N (\bm{x})
\end{bmatrix}
+
\begin{bmatrix}
\epsilon_1 (\bm{x})\\ \epsilon_2(\bm{x})\\ \vdots \\ \epsilon_N (\bm{x})
\end{bmatrix}
=F(\bm{x}) + E(\bm{x}),
\end{gathered}
\end{equation*}

\noindent
where $F:\mathcal{R}^\mathcal{D} \rightarrow \mathcal{R}^N $ is zero mean multivariate process with covariance $\mbox{cov}^f_{ij}(\bm x,{\bm x}^{\prime})$ $=\mbox{cov}^f_{ij} \big ( f_i(\bm x),f_j({\bm x}^{\prime}) \big)$ for $i,j \in \mathcal{I}$ and $\epsilon_i(\bm x) \sim \mathcal{N}(0,\sigma^2_i)$ represents additive noise. For the $i$th output the observed data is denoted as $\mathcal{D}_i=\{(\bm{y}_i,\bm{X}_i)\}$, where $\bm y_i = [y_i^1,\cdots,y_i^{p_i}]^{\top}$, $y_i^c:=y_i(\bm{x}_{ic})$, $\bm{X}_i=[\bm{x}_{i1},\cdots,\bm{x}_{ip_i}]^{\top}$ and $p_i$ represents the number of observations for output $i$. Now let $P=\sum p_i$ and $D_{\mathcal{I}} =\{D_1, \cdots, D_N\}$, then the predictive distribution for output $i$ at a new input point $\bm{x}_0$ is given as
\begin{align}
\label{eq:predictive}
pr(y_i(\bm{x}_0)|D_{\mathcal{I}})&=\mathcal{N}  \big(    \bm{C}^{\top}_{\bm f,f_i^0} (\bm{C}_{\bm f,\bm f} + \bm \Sigma)^{-1}\bm{y},\ C_{f_i^0,f_i^0}+ \notag
\\ & \sigma^2_i-\bm{C}^{\top}_{\bm f,f_i^0} (\bm{C}_{\bm f,\bm f} + \bm \Sigma)^{-1}\bm{C}_{\bm f,f_i^0} \big) ,
\end{align} 
\noindent
where $\bm{y}=[\bm{y}_1^{\top},\cdots,\bm{y}_N^{\top}]^{\top}$ corresponding to the latent function values $\bm{f}=[\bm{f}_1^{\top},\bm{f}_2^{\top},...,\bm{f} _N^{\top}]^{\top}$, $f_i^c:=f_i(\bm{x}_{ic})$ such that $\bm{f}_i=f_i(\bm{X}_i)$, $\bm{C}_{\bm f,\bm f} \in \mathcal{R}^{P\times P}$ is the covariance matrix from the operator $\mbox{cov}^f_{ij}(\bm x,{\bm x}^{\prime})$ and $\bm \Sigma=\operatorname{diag}[\sigma^2_1 \bm{I}_{p_1},...,\sigma^2_N\bm{I}_{p_N} ]$ is a block diagonal matrix with $\bm{I}$ as the identity matrix. Finally, $C_{f_i^0,f_i^0}=\mbox{cov}^f_{ii}(\bm{x}_0,{\bm{x}_0})$, where $f_i^0:=f_i(\bm{x}_0)$ and $\bm{C}_{\bm f,f_i^0}=[\bm{C}^{\top}_{\bm{f}_1,f_i^0},\cdots,\bm{C}^{\top}_{\bm{f}_N,f_i^0}]^{\top}$; $\bm{C}_{\bm{f}_c,f_i^0}=[\mbox{cov}^f_{ic}(\bm{x}_0,\bm{x}_{c1}),\cdots,\mbox{cov}^f_{ic}(\bm{x}_0,\bm{x}_{cp_c})]^{\top}$.

As shown in (\ref{eq:predictive}), information transfer is facilitated via $\mbox{cov}^f_{ij}(\bm x,{\bm x}^{\prime})$. Under the $\mathcal{CP}$ in (\ref{eq:cp}) and assuming independent latent function $X_q$ with  $\mbox{cov}(X_i(\bm{u}),X_i(\bm{u}^{\prime}))=\delta(\bm{u}-\bm{u}^{\prime} )=\delta_{\bm{u}\bm{u}^{\prime}}$ ($\delta$ is the Dirac delta function) we have 
\begin{align}
\label{eq:cov}
\mbox{cov}^f_{ij}(\bm x,{\bm x}^{\prime})=\sum_{q=1}^{Q} \int_{-\infty}^{\infty}K_{qi}(\bm{u}) K_{qj}(\bm{u}-\bm{d}) d \bm{u}.
\end{align}
where $\bm{d}=\bm{x}-{\bm x}^{\prime} \in \mathcal{R}^D$ denotes a convolution. Here we note that a more general case can be used where $X_i$ is a $\mathcal{GP}$ generated from a $\mathcal{CP}$, i.e.,  $\mbox{cov}(X_i(\bm{u}),X_i(\bm{u}^{\prime}))=\int K_{X_i}(\bm{u})K_{X_i}(\bm{u}-\bm{d})d\bm{u}$. In the appendix we show that the following results also hold under such a case. 

\section{Negative Transfer: Definition and Conditions} \label{sec:define}
In this section, we first give a strict definition of negative transfer in $\mathcal{MGP}$, and then explore necessary conditions needed to avoid negative transfer.
 
\subsection{Definition of negative transfer}
Similar to multi-task learning, negative transfer draws its roots from transfer learning \citep{pan2009survey}. A widely accepted description of negative transfer is stated as \emph{``transferring knowledge from the source can have a negative impact on the target learner''}. In an $\mathcal{MGP}$, negative transfer could be defined similarly: the integrative analysis of all outputs can have negative impact on the performance of the model compared with separate modeling of each output or a subset of them.


\begin{definition}\label{def:negative1}
	Consider an $\mathcal{MGP}$ with $N$ possible outputs, and assume $y_i$ represents the target output. Let the index set of all outputs $\mathcal{I}$ comprise of $M$ non-empty disjoint subsets $\mathcal{I}=\{\mathcal{I}_1 \cup \cdots  \mathcal{I}_m \cup \cdots \mathcal{I}_M\}$.  Then, we can define the information transfer metric ($IT$) of the $i^{th}$ output $y_i$, $i\in \mathcal{I}_m$ as follows:
	$$IT_i(D_{\mathcal{I}_m})=R_i[\mathcal{MGP}(D_{\mathcal{I}})]- R_i[\mathcal{MGP}(D_{\mathcal{I}_m})] , $$	 
	where $\mathcal{MGP}(D_u)$ is a $\mathcal{MGP}$ using data $D_u$, $R_i[\mathcal{MGP}(D_u)]=E\big[\mathcal{L}\big(y_i(\bm{x}),y_{i, true}(\bm{x})\big)\big|D_u]$ defines the expected risk using some loss function $\mathcal{L}$ and $y_i(\bm{x})$ denotes the predicted random variable in (\ref{eq:predictive}). We say negative transfer occurs if $IT_i(D_{\mathcal{I}_m})$ is positive.
\end{definition}

Definition \ref{def:negative1} implies that negative transfer happens when using $D_\mathcal{I}$ leads to worse accuracy compared to using a subset of the data or just an individual $\mathcal{GP}$. Therefore, one can provide a model flexible enough to avoid negative transfer if there exists an $\mathcal{MGP}$ such that $\forall \> \bm{x}_0 \in \mathcal{R}^\mathcal{D}$ 
\begin{align} \label{eq:maincondition1}
pr(y_i(\bm{x}_0)|D_{\mathcal{I}})=pr(y_i(\bm{x}_0)|D_{\mathcal{I}_m}), \forall \> i \in \mathcal{I}_m \subseteq \mathcal{I}  
\end{align}
One can think of $\mathcal{I}_m$ as the index for the subset of outputs that share commonalities with $y_i$ ($\mathcal{I}_m$ here includes $i$). For instance, if $y_i$ shares no commonalities with any other output ($\mathcal{I}_m=i$) then the $\mathcal{MGP}$ should be able to have $pr(y_i(\bm{x}_0)|D_{\mathcal{I}})=pr(y_i(\bm{x}_0)|D_i)$ for all $\bm{x}_0$, i.e., the conditional predictive distribution in (\ref{eq:predictive}) for output $i$ is independent of all other outputs. In other words, we need an $\mathcal{MGP}$ that is able to collapse to independent $\mathcal{GP}s$ or an $\mathcal{MGP}$ with only related outputs. 

Building a model that can achieve (\ref{eq:maincondition1}) is a challenging task since, following the conditional predictive distribution in (\ref{eq:predictive}), (\ref{eq:maincondition1}) occurs if and only if \citep{whittaker2009graphical} $ \forall \>  \bm{x},\bm{x}^{\prime} \in \mathcal{R}^\mathcal{D}$
\begin{align} \label{eq:maincondition2}
\mbox{cov}^f_{ij}(\bm x,{\bm x}^{\prime})=0, \forall \> i \in \mathcal{I}_m \>\text{and}\> j \in \mathcal{I}_{/\mathcal{I}_m}
\end{align}
In the following section we study the necessary condition for the $\mathcal{MGP}$ to achieve (\ref{eq:maincondition2}). Specifically we show that if $M$ is known (i.e., we know how many distinct/unrelated subgroups of outputs exist) then to achieve (\ref{eq:maincondition2}) the necessary condition is that $Q \geq M$. However the fact that $M$ is not known before hand implies that we need $Q \geq N$.

\subsection{Conditions to avoid negative transfer} \label{sec:conditions}
We first provide a lemma based on the $\mathcal{CP}$ covariance in (\ref{eq:cov}) needed to establish our result. 

\begin{lemma} \label{lemma1}
	Given two outputs, $y_1$ and $y_2$, modeled using one latent function $X_1$. Assume, the kernels $K_{1i}(\bm{x}) \in L^1(\mathcal{R}^\mathcal{D})$, $i \in \{1,2\}$, satisfy one of the following conditions:	
	\begin{itemize}
		\item $K_{1i}(\bm{x})=\alpha_{1i} k_{1i}(\bm{x})$, $\alpha_{1i} \in \mathcal{R}$ and $k_{1i}(\bm{x}) >0 \> \forall \> \bm{x} \in \mathcal{R}^\mathcal{D}$. Typical cases include squared exponential, Matern, quadratic kernel, periodic and local periodic.
		\item $k_{1i}$ has the form $\sum_u {a_u^2} \text{exp}\big(\bm{x}^T\bm{B}_u\bm{x}\big)\text{cos}(2\pi{\bm{c}_u}^T\bm{x})$ with parameters $({a_u}, \bm{B}_u,{\bm{c}_u})$. Typical cases include the Spectral, generalized spectral, MOCSM \citep{chen2019multioutput}, CSM \citep{ulrich2015gp} and SMD \citep{chen2018spectral} kernels.
	\end{itemize} 
Then for $\forall \bm{x}, {\bm x}^{\prime} \in \mathcal{R}^\mathcal{D}$,
$\mbox{cov}^f_{12}(\bm x,{\bm x}^{\prime})=\int_{-\infty}^{\infty}K_{11}(\bm{u}) K_{12}(\bm{u}-\bm{d}) d \bm{u}=0$ if and only if at least one of $K_{11}$ and $K_{12}$  is identically equal to zero. 
\end{lemma} 

The technical details for Lemma \ref{lemma1} are given in the Appendix A. Clearly when using one latent function $X_1$ if one of the kernels is identically zero then the $\mathcal{MGP}$ is invalid ($\mbox{cov}^f_{uu}(\bm x,{\bm x}^{\prime})=0 \> \forall \> \bm{x}, u \in \{1,2\}$). On the other hand, if we use $Q \geq 2$ latent functions, then the model has enough flexibility to construct $f_i$ from different latent functions, i.e. $f_1 = K_1\star X_1$ and $f_2 = K_2\star X_2$. In this case, $\mbox{cov}^f_{12}=0,\ \forall \> \bm{x}, {\bm x}^{\prime}$. Hence, Lemma \ref{lemma1} implies that only if $Q \geq 2$ we can achieve $\mbox{cov}^f_{12}(\bm x,{\bm x}^{\prime})=0 \> \forall \> \bm{x}, {\bm x}^{\prime}$.

In Lemma \ref{lemma1} the assumption that kernels belong to the $L^1$ space is also needed for a stable $\mathcal{CP}$ construction in (\ref{eq:cp}). Here we note that despite the fact that the conditions presented  satisfy most (if not all) of the kernels currently used in the $\mathcal{CP}$, in the appendix we also provide some simple means to check the conditions. For instance, for any even kernel $K(\bm{x})=K(-\bm{x})$, then $\mbox{cov}^f_{12}(\bm x,{\bm x}^{\prime})=K_{11} \star K_{12}$. Hence, since the Fourier operator $\mathcal{F}$ is injective, it suffices to show that $\mathcal{F}\{K_{11}\}(\xi)\cdot\mathcal{F}\{K_{12}\}(\xi)=0$ if and only if $\mathcal{F}\{K_{11}\}$ or $\mathcal{F}\{K_{12}\}$ is identically zero. 

We now give the main theorem for the necessary condition to avoid negative transfer.

\begin{theorem}\label{thm1}
	Given $K_{1i}(\bm{x}) \in L^1(\mathcal{R}^\mathcal{D})$ that satisfies the conditions in Lemma \ref{lemma1} then there exists an $\mathcal{MGP}$, constructed using a $\mathcal{CP}$, that can achieve $pr(y_i(\bm{x}_0)|D_{\mathcal{I}})=pr(y_i(\bm{x}_0)|D_{\mathcal{I}_m}), \forall \> i \in \mathcal{I}_m$ if and only if we have $Q \geq M$ latent functions.
\end{theorem}

The technical details for Theorem \ref{thm1} are given in the Appendix B. 
It is crucial to note here that in reality we do not know $M$, i.e., we do not know how many distinct subgroups that are uncorrelated exist. Thus, in order to guarantee that negative transfer can be avoided we need $Q \geq M=N$. This also implies that the model is flexible enough to collapse to $N$ independent $\mathcal{GP}$s and hence predict each output independently. An illustrative example is also given in Appendix C.  

\subsection{Induced Challenges}
Despite the many works in the previous decade on reducing the computational complexity of both the $\mathcal{GP}$ and $\mathcal{MGP}$, the results in Sec.\ref{sec:conditions} induce another key challenge for $\mathcal{MGP}$: the\textit{ high dimensional parameter space}. This challenge is inherited from the $\mathcal{CP}$ construction which provides different covariance parameters (via the kernels) to different outputs levels. For instance, assume any kernel $K_{qi}(\bm{x})$ has $\omega$ parameters to be estimated then using the $\mathcal{CP}$, this implies estimating $QN\omega+N$ parameters, where the added $N$ parameters are for $\epsilon_i(\bm{x})$. Following our results, a model that can avoid negative transfer thus needs at least $N^2\omega+N$. Note here that $\omega$ also increases with $\mathcal{D}$, i.e., the dimension of $\bm{x}$. Obtaining good estimates in such a high dimensional space is an impractical task specifically under a non-convex and highly nonlinear objective, be it the exact Gaussian likelihood or its variational bound. \textit{Indeed, it is crucial to note that computational complexity and parameter space are two separate challenges and the many papers that tackle the former still suffer from the latter challenge}. We conjecture that for this reason, most (if not all) $\mathcal{MGP}$ literature (including all aforementioned cited papers) have used $1 \leq Q \leq 4$. 

To address this challenge, in Sec. \ref{sec:relaxation} we provide relaxation models that can significantly reduce the parameters space and scale to arbitrarily large datasets by parallelization. Further, our proposed relaxations allow any sparse approximation to be directly plugged in. This in turns allows utilization of the many advances in reducing the computational complexity (inducing point, state space approximation, etc..). 

\section{Relaxation models}\label{sec:relaxation}
In this section we propose two relaxation models: The arrowhead and pairwise model. Without loss of generality, we focus on predicting output $y_1$ using the other $N-1$ outputs. We use $\mathcal{I}_{/1}$ to index all outputs except $y_1$.

\subsection{Arrowhead Model}
\begin{figure}[!htbp]
	\centering
	\makebox[\linewidth]{\includegraphics[keepaspectratio=true,width=\linewidth]{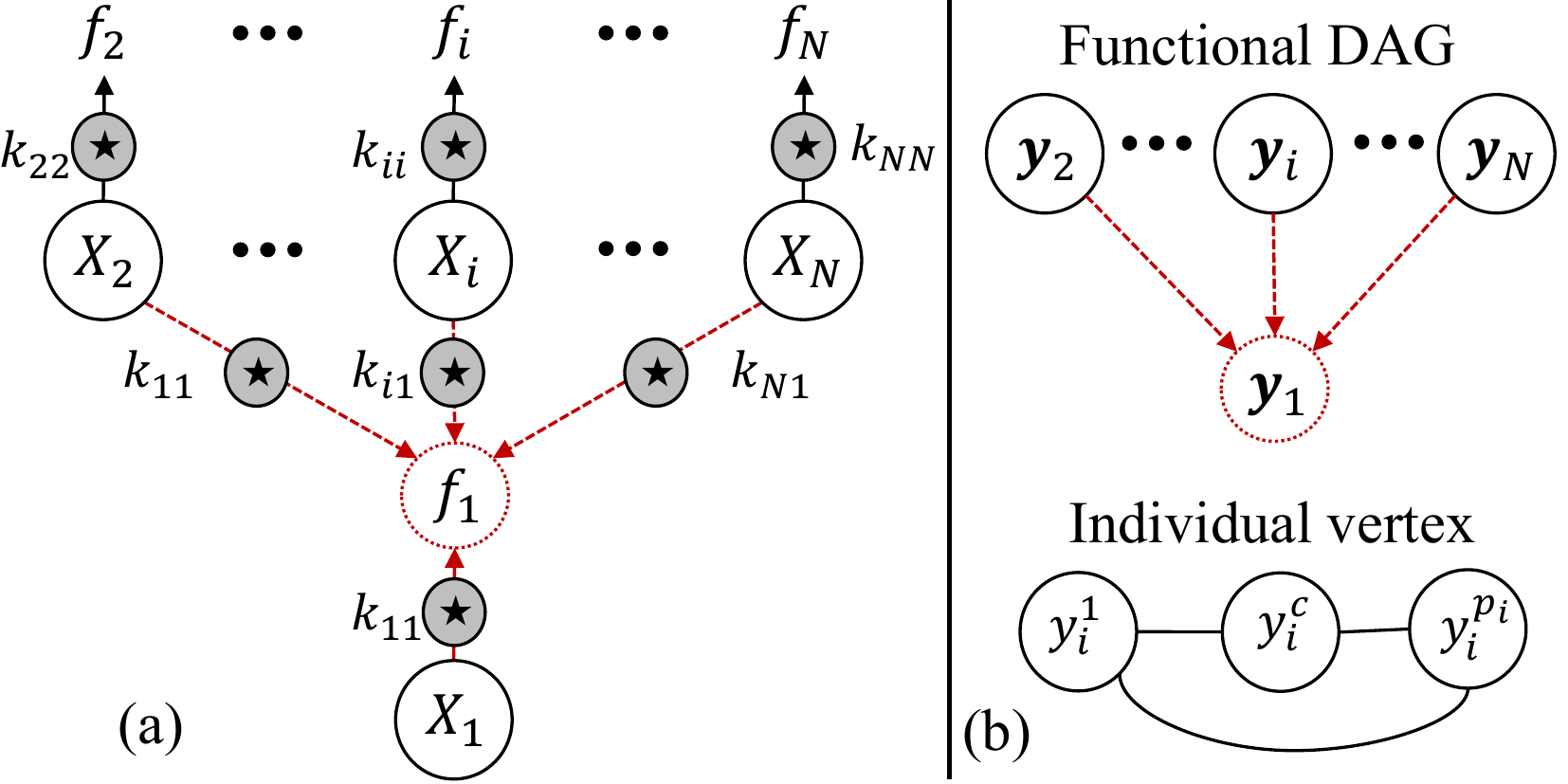}}
	\caption{Arrowhead model}
	\label{fig:arrowhead}
\end{figure}

The idea of an arrowhead model is originated from the arrowhead matrix. While still using $N$ latent functions, we can assume that all outputs $y_i$, $i \in \mathcal{I}_{/1}$, are independent and only share information with $y_1$, the output of interest. This implies that $\mbox{cov}^f_{ij}(\bm x,{\bm x}^{\prime})=0, \forall \> i,j \in \mathcal{I}_{/1}$. The structure and covariance matrix are highlighted in Fig. \ref{fig:arrowhead}(a) and (\ref{eq:cov_arrow}) respectively. As shown in the figure, $y_1$ possesses unique features encoded in $X_1$ and shared features with other outputs encoded in $X_i, i\in \mathcal{I}_{/1}$.
\begin{align}\label{eq:cov_arrow}
\bm{C}_{\bm{f},\bm{f}}^{P \times P}=
\renewcommand\arraystretch{1.2}
\begin{pmatrix} \begin{array}{c|ccc}
\bm{C}_{\bm{f}_1,\bm{f}_1} & \bm{C}_{\bm{f}_1,\bm{f}_2}& \dots & \bm{C}_{\bm{f}_1,\bm{f}_N} \\ \hline 
\bm{C}_{\bm{f}_2,\bm{f}_1} & \bm{C}_{\bm{f}_2,\bm{f}_2}& \dots & \bm{0}_{p_2\times p_N} \\ 
\vdots  &\vdots& \ddots & \vdots \\
\bm{C}_{\bm{f}_N,\bm{f}_1} &\bm{0}_{p_N\times p_2}& \dots & \bm{C}_{\bm{f}_N,\bm{f}_N}
\end{array}\end{pmatrix}
\end{align}

The arrowhead structure in fact poses many unique advantages: (1) Linear increase of parameter space dimension with $N$: The number of parameters to be estimated is reduced to $(2N-1)\omega + N$. (2) provides enough flexibility to achieve (\ref{eq:maincondition2}) and hence avoid negative transfer. For instance, if $\mbox{cov}^f_{1i}(\bm x,{\bm x}^{\prime})=0 \> \forall \bm{x}$ and $i \in \mathcal{I}_{/1}$ then outputs $y_1$ and $y_i$ are predicted independently. (3) can be parallelized, where independent arrowhead models are build to predict each output. (4) One nice interpretation of the arrowhead structure is through a Gaussian directed acyclic graphical model (DAG) with vertices $V=\{\bm{y}_i:i\in I\}$. Unlike typical DAGs, each vertex in this graph is in itself a fully connected undirected Gaussian graphical model, i.e. a functional response. This is shown in Fig.\ref{fig:arrowhead}(b). Based on this, the full likelihood factorizes over parent nodes. To see this, let $\mathcal{L}(\bm{\theta};\bm{y})$ denote the likelihood of the dataset, where $\bm{\theta}=\{\bm{\theta}^{\top}_f,\bm{\sigma}^{\top}\}^{\top}$, such that $\bm{\theta}_f$ and $\bm{\sigma}$ are kernel and noise parameters. Then, $\mathcal{L}(\bm{\theta};\bm{y})=\mathcal{L}^{1|i \in \mathcal{I}_{/1}}(\bm{\theta};\bm{y}_1|\bm{y}_2, \cdots, \bm{y}_N) \prod_{i \in \mathcal{I}_{/1}}\mathcal{L}^i(\bm{\theta};\bm{y}_i)$. This reduces the complexity of exact inference to $O(Np^3)$ assuming $p_i=p \> \forall i$, i.e, complexity of $N$ independent $\mathcal{GP}$s. This complexity is similar to the well known inducing point sparse approximation in \citet{alvarez2009sparse}, however without the assumption of conditional independence given discrete observations from the latent functions.  

Despite reduced complexity, the main advantge is the reduction in the parameter space. Here it is crucial to note that any sparse approximation, be it an inducing point/variational approximation, a state space approximation, a matrix tapering approach or just a faster matrix inversion/determinant calculation scheme, can be plugged in into this structure. 
\subsubsection{Encouraging sparsity via regularization} \label{sec:penalty1}
Besides the fact that the arrowhead model can avoid negative transfer, an interesting feature is that we can add regularization that helps reduce negative transfer and is capable of automatic variable selection. Here variable selection implies selection of which functions should be predicted independently or not. To see this, let $\ell(\bm{\theta};\bm{y})=-\mbox{log} \ \mathcal{L}(\bm{\theta};\bm{y})= \frac{1}{2} \langle \bm{Y},(\bm{C}_{\bm f,\bm f} + \bm \Sigma)^{-1}\rangle + \frac{1}{2}\mbox{log}|\bm{C}_{\bm f,\bm f} + \bm \Sigma| $ where $\langle \bm{A} , \bm{A}^{\prime} \rangle=\operatorname{trace}(\bm{A} \bm{A}^{\prime})$ and $\bm{Y}=\bm{y}\bm{y}^{\top}$. A penalized version of $\ell(\bm{\theta};\bm{y})$  is defined as 
\begin{align}
\label{eq:penalty}
\ell_{\mathbb{P}}(\bm{\theta};\bm{y},\lambda)=\ell(\bm{\theta};\bm{y})+\mathbb{P}_\lambda(\bm{\theta}_0) \ ,
\end{align}
\noindent
where $\mathbb{P}_\lambda(|\bm{\theta}_0|)$ is a penalty function and $\bm{\theta}_0 \subseteq \bm{\theta}$. Possible choices of $\mathbb{P}_\lambda$ and statistical guarantees on (\ref{eq:penalty}) are provided later in Sec. \ref{sec:penalty}. One can directly observe that for $K_{qi}(\bm{x})=\alpha_{qi} k_{qi}(\bm{x})$, then $\bm{\theta}_0=\{\alpha_{q1}\}_{q=2}^{N}$. Therefore when $\alpha_{q1} \rightarrow 0$
\begin{align}
&\mbox{cov}^f_{1i}(\bm x,{\bm x}^{\prime})=\int_{-\infty}^{\infty}K_{i1}(\bm{u}) K_{ii}(\bm{u}-\bm{d}) d \bm{u} \rightarrow 0 \notag.
\end{align}
and hence outputs $y_1$ and $y_{q=i}$ will be predicted independently. Thus any shrinkage penalty will encourage the arrowhead model to limit information sharing across unrelated output. Another advantage besides automatic shrinkage is \textit{functional variable selection} where the sparse elements in $\{\alpha_{q1}\}_{q=1}^{N}$ would identify which outputs are related to $y_1$.     

\subsection{Pairwise Model}
Despite the linear increase in parameter space in the arrowhead model, when $N$ is extremely large model estimation can still be prohibitive. To this end, and inspired by the work of \citet{fieuws2006pairwise}, we propose distributing the  $\mathcal{MGP}$ into a group of bivariate $\mathcal{GP}$s which are independently built. Predictions are then obtained through combining predictions from each bivariate $\mathcal{GP}$. As previously mentioned we focus on predicting output 1 through borrowing strengths from the other $N-1$ outputs. 

\begin{figure}[!htbp]
	\centering
	\makebox[\linewidth]{\includegraphics[keepaspectratio=true,width=\linewidth]{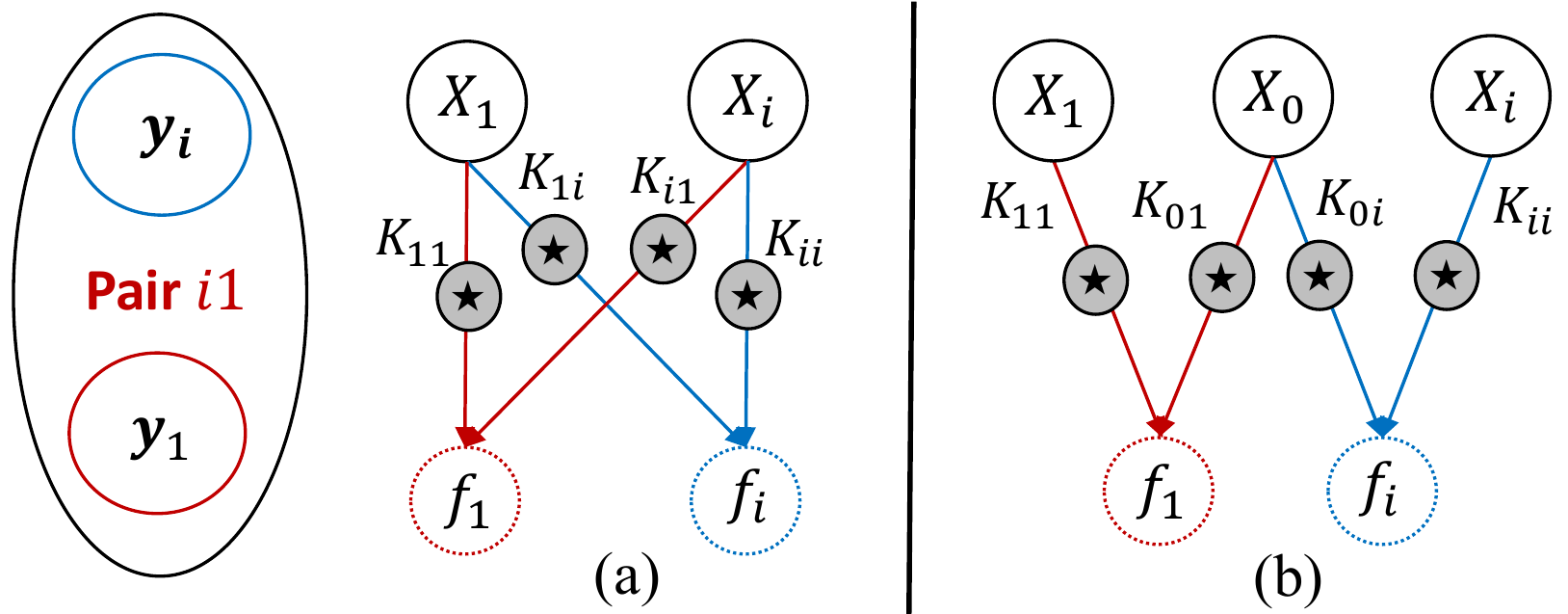}}
	\caption{pairwise model}
	\label{fig:pairfig}
\end{figure}

Fig. \ref{fig:pairfig}(a) illustrates the pairwise submodel between $y_1$ and $y_i, i\in\mathcal{I}_{/1}$, where two latent functions are used to avoid negative transfer. Note here that the structure in Fig. $\ref{fig:pairfig}$(b) is proposed for efficient regularization and is discussed later in Sec. \ref{sec:penalty2}. The key advantage of the pairwise structure is that: (1) it can scale to an arbitrarily large $N$ by parallelization where each submodel is estimated with a limited number of parameters ($4\omega+2$) and with complexity of $O(2p^3)$ (assuming exact inference with no approximations and that $p_i=p \> \forall i$) (2) It can avoid negative transfer without the need for $Q=N$ latent functions. 

After building the $N-1$ sub-models, combining predictions boils down to combining $N-1$ predictive distributions $pr(y_1(\bm{x}_0)|D_1, D_i)_{i \in \mathcal{I}_{/1}}$ in (\ref{eq:predictive}). This can be readily done using the rich literature on product of experts and Bayesian committee machines (see \citet{deisenroth2015distributed},  \citet{moore2015gaussian} and \citet{tresp2000bayesian} for an overview). The key idea across such approaches, in our context, is that sub-models that are uncertain about their predictions of $y_1(\bm{x}_0)$  (i.e., have larger predictive variance) will get less weight. 

\subsubsection{Encouraging sparsity in pairwise model} \label{sec:penalty2}
Similar to the arrowhead model, the pairwise approach also facilitates regularization and automatic variable selection. For the structure illustrated in Fig. $\ref{fig:pairfig}$(a), unlike (\ref{eq:penalty}), we have that $ 
\mbox{cov}^f_{1i}(\bm x,{\bm x}^{\prime})=\int_{-\infty}^{\infty}K_{11}(\bm{u}) K_{1i}(\bm{u}-\bm{d}) d \bm{u} + \int_{-\infty}^{\infty}K_{i1}(\bm{u}) K_{ii}(\bm{u}-\bm{d}) d \bm{u}$. Therefore to encourage sparsity, a group penalty $\mathbb{P}^{G}_\lambda$ on $K_{i1}$ and $K_{1i}$ is needed. 
\begin{align}
\label{eq:penalty_pair}
\ell_{\mathbb{P}}(\bm{\theta}_{1i};\bm{y}_1, \bm{y}_i,\lambda)=\ell(\bm{\theta}_{1i};\bm{y}_1, \bm{y}_i)+\mathbb{P}^{G}_\lambda(\alpha_{1i},\alpha_{i1}).
\end{align}
One well-known option for $\mathbb{P}^{G}_\lambda$ is the group Lasso $\mathbb{P}^{G}_\lambda=\sqrt{2}\lambda||({\alpha_{1i},\alpha_{i1}})^T||_2$. 

Alternatively, one can utilize the structure in Fig. $\ref{fig:pairfig}$(b) and instead of penalizing the kernels, one can regularize the shared latent function $X_0$. For instance, one can augment the covaraince of $X_0$ with a parameter $\alpha_0$ such that $\mbox{cov}(X_0(\bm{u}),X_0(\bm{u}^{\prime}))=\alpha_0 \delta(\bm{u}-\bm{u}^{\prime} )$. Then, 
\begin{align}
\label{eq:penalty_pair2}
\ell_{\mathbb{P}}(\bm{\theta}_{1i};\bm{y}_1, \bm{y}_i,\lambda)=\ell(\bm{\theta}_{1i};\bm{y}_1, \bm{y}_i)+\mathbb{P}^{G}_\lambda(\alpha_{0}).
\end{align}
It can be directly verified that as $\alpha_0 \rightarrow 0$ outputs $y_1$ and $y_i$ are predicted independently. 

\section{Guarantee on Penalization} \label{sec:penalty}
In this section we establish a statistical guarantee on the estimates $\hat{\bm{\theta}}$, under the penalized setting in (\ref{eq:penalty}). Our result extends the well known consistency for independent and identically normal data to the $\mathcal{MGP}$ case with correlated data. Prior to that, we first briefly discuss different forms of  $\mathbb{P}_\lambda(\bm{\theta}_0)=\sum_i \mathbb{P}_\lambda(|\theta_{i}|) $. Possible well-known choices include the ridge penalty $\mathbb{P}_\lambda(|\theta_i|)=\lambda \theta_i^2$, $L^1$ penalty $\mathbb{P}_\lambda(|\theta_i|)=\lambda |\theta_i|$, bridge penalty $\mathbb{P}_\lambda(|\theta_i|)=\lambda |\theta_i|^{0 <  \cdot < 1}$, and SCAD penalty which includes two tuning parameters ($\lambda$ and $\gamma$) $\mathbb{P}_\lambda(|\theta_i|)=\lambda |\theta_i|$ if $|\theta_i| \leq \lambda$, $(\theta_i^2-2\gamma\lambda|\theta_i|+\lambda^2)/(2\gamma-2)$ if $\lambda<|\theta_i| \leq \gamma\lambda$, $\lambda^2(\gamma+1)/2$ if $|\theta_i| > \gamma\lambda$ \citep{fan2001variable}. The tuning parameters can be estimated using cross validation \cite{friedman2001elements}. Next we provide the main theorem. 

Assume that $\bm{\theta}^{\ast}$ corresponds to the true parameters,  \raisebox{0.3ex}{``$\prime$"} denotes a derivative and $r_P$ is a sequence such that $r_P \rightarrow \infty$ as $P \rightarrow \infty$, then we have that 

\begin{theorem}\label{thm2}
Given that $\mathbb{P}_\lambda(|\theta_i|)\geq 0 $, $\mathbb{P}_\lambda(0)= 0$ and $\mathbb{P}_\lambda(|\theta_i^{\prime}|) \geq \mathbb{P}_\lambda(|\theta_i|)$ if $|\theta_i^{\prime}| \geq |\theta_i|$.  If $\mbox{max}\{|\mathbb{P}^{\prime\prime}_\lambda(|\theta_i^{\ast}|)|:\theta_i^{\ast} \neq 0\} \rightarrow 0$, then $\exists$ $\hat{\bm{\theta}}$ in $\ell_{\mathbb{P}}(\bm{\theta};\bm{y},\lambda)$, such that $||\hat{\bm{\theta}}-\bm{\theta}^{\ast}||=O(r^{-1}_P+r)$, where $r=\mbox{max}\{\mathbb{P}^{\prime}_\lambda(|\theta_i^{\ast}|):\theta_i^{\ast} \neq 0\}$. 
\end{theorem}
Theorem \ref{thm2} shows that for the penalized likelihood, the true parameter estimates would still be asymptotically retained. This provide theoretical justification for our regularization approach in both the arrowhead and pairwise models. Technical details and further discussions are deferred to Appendix D.
\section{Proof of Concept and Case Studies} \label{sec:case}
Since negative transfer is a subject yet to be explored in $\mathcal{MGP}$, we dedicate most of this section towards  a proof of concept for: (1) the impact of negative transfer, (2) the need for sufficient latent functions as shown in theorem \ref{thm1}, (3) the advantageous properties of the proposed latent structures, (4) the role of regularization in selection of related outputs as shown in theorem \ref{thm2}. Two case studies are then presented, while additional studies and numerical examples are differed to Appendix G.
\subsection{Illustration of Negative Transfer}
\subsubsection{Convolved Squared exponential Kernel} \label{sim:exp}
In this setting, we aim to illustrate theorem \ref{thm1} using the well-known convolved squared exponential kernel in \citet{alvarez2011computationally}. We generate outputs $y_1$, $y_2$ and $y_3$ from 
\begin{eqnarray*}
	 y_1(x) &=&  5\cdot\sin(3x/2)+\epsilon_1(x) \\
    y_2(x) &=&  5\cdot \sin(x)-3+\epsilon_2(x) \\
    y_3(x) &=&  x^2/10-5+\epsilon_3(x)
\end{eqnarray*}
where $x\in \mathcal{R}$ is evenly spaced in $[0,10]$, $p_1=p_2 = p_3=20$ and $\sigma_1=\sigma_2=\sigma_3=0.05$. In Table \ref{tb:table_square} we report the means squared error (MSE), averaged over the 3 outputs, on $p=70$ uniformly spaced points in $[0,10]$ when $Q=1,2,3$ and $4$. Table \ref{tb:table_square} provides many interesting insights. Indeed from the function specifications, it is clear that they have very different shape and length scales (i.e., frequency and amplitude). As a result, when using one or two latent functions negative transfer leads to large predictive errors. It is also noticeable that the result of using $Q=4$ does not have much difference with that of $Q=3$. This confirms our theorem which implies that with at least $N$ latent functions an $\mathcal{MGP}$ is capable of avoiding negative transfer. 
\begin{table}[!htb] 
	\caption{Predictive error with varying $Q$}
	\label{tb:table_square}
	\centering
		\begin{tabular}{ccccc}
			\toprule 
			Q& 1& 2& 3& 4\\
			\midrule
			MSE & 25.183& 11.464& 0.00159&  0.00157\\
			\bottomrule 
		\end{tabular}
\end{table}

\subsubsection{Spectral Kernel}
The immediate follow up question is what if we use the recently proposed, more flexible class of spectral kernels. The aim is to illustrate that as shown in Lemma \ref{lemma1}, avoiding negative transfer is mainly independent of what kind of kernel we use, i.e. even if we use a more flexible kernel. We use the same data with that in setting \ref{sim:exp}. The covariance function is given as:
$$\operatorname{cov}^f_{ij}(x,x^{\prime})=\sum_{q=1}^Q \dfrac{a_{qi}a_{qj}}{2}\sqrt{\dfrac{\pi}{\sigma_{qi}^2+\sigma_{qj}^2}}H(d)$$

where $H(d)=e^{A_{1}(d)} \cos \left(\theta_{1} d\right)+e^{A_{2}(d)} \cos \left(\theta_{2} d\right)$. Formulation of $A_{i}(d)$ and $\theta_i$, $i=1,2$ are given in Appendix E. This covariance is the result of a convolution across spectral kernels.

\begin{figure}[!htbp]
	\centering
	\makebox[\linewidth]{\includegraphics[keepaspectratio=true,width=\linewidth]{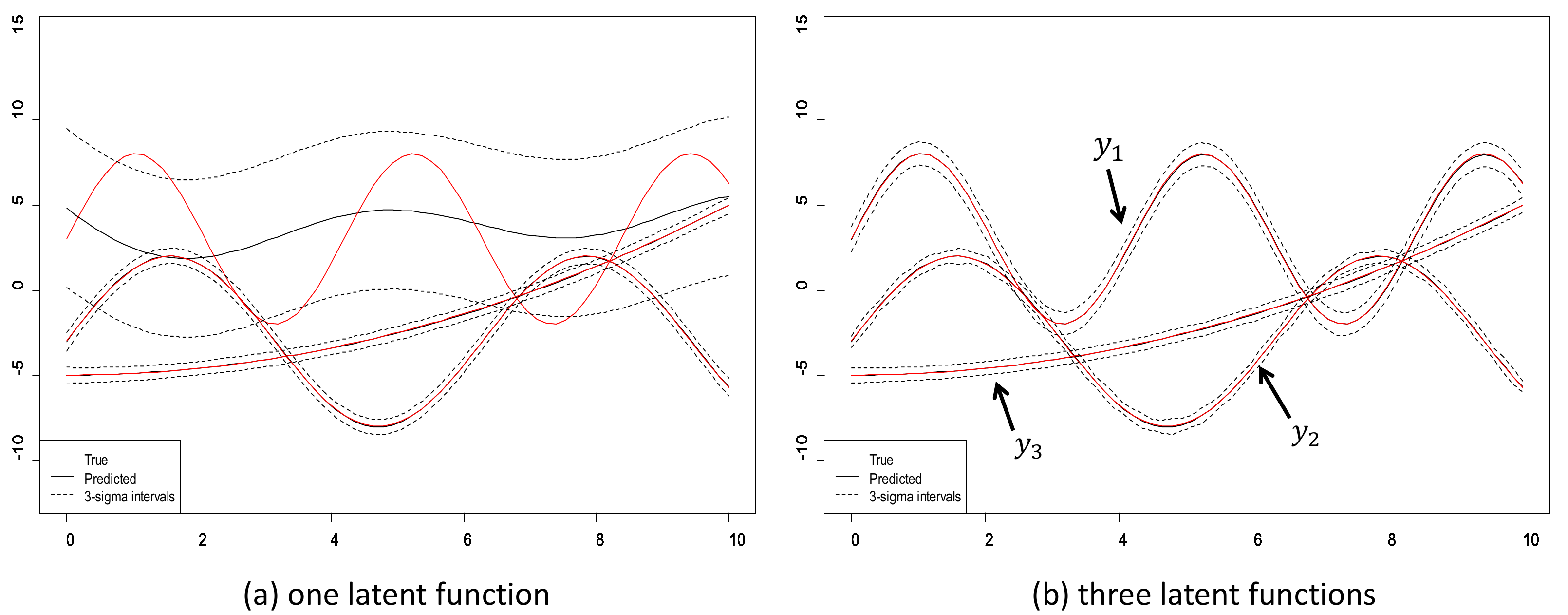}}
	\caption{Illustration of predictions using a spectral kernel}
	\label{spectral}
\end{figure}

The predictive results for the three outputs are illustrated in Fig. \ref{spectral}. The results confirm that even with a flexible kernel negative transfer will detrimentally effect model performance without enough latent functions. Indeed in Fig. \ref{spectral}(a) one can observe that $y_1(x)$ and $y_3(x)$ have larger length scales and hence are smoother. As a results when $Q=1$, output $y_2(x)$ is forced to have a larger length scale in lieu of the two other outputs. This however, can be avoided with a sufficient number of latent functions as shown in Fig. \ref{spectral}(b).

%
%
%
%

\subsection{Role of Regularization}
Still under the setting of Sec \ref{sim:exp}, we try to verify theorem \ref{thm2} and the impact of $\mathbb{P}_\lambda$ on automatic selection of related output. We use the pairwise model described in Fig. \ref{fig:pairfig}(b) where two bivariate submodels are used to predict $y_1$: $(y_1,y_2)$ and $(y_1, y_3)$. The covariance function of $y_1$ and $y_i$ ($i=2,3$) constructed using Fig. \ref{fig:pairfig}(b) are given as 
\begin{align*}
\operatorname{cov}_{11}^f(d)&= \alpha_{11}^2\exp\{-\dfrac{d^2}{4\cdot l_{11}^2}\}+ \alpha_{01}^2\exp\{-\dfrac{d^2}{4\cdot l_{01}^2}\}+\sigma^2\\
\operatorname{cov}_{ii}^f(d) &= \alpha_{ii}^2\exp\{-\dfrac{d^2}{4\cdot l_{ii}^2}\}+ \alpha_{0i}^2\exp\{-\dfrac{d^2}{4\cdot l_{0i}^2}\}+\sigma^2\\
\operatorname{cov}_{1i}^f(d) &= \alpha_{01}\alpha_{0i} \sqrt{\dfrac{2|l_{01}l_{0i}|}{l_{01}^2+l_{0i}^2}}\exp\{-\dfrac{1}{2}\dfrac{(d-\mu_{1i}^2)}{l_{01}^2+l_{0i}^2}\}
\end{align*}

We applied the pairwise model with a regularization term respectively to the data. For the penalty we use
$\mathbb{P}_\lambda^G(\bm{\alpha}_0)=\lambda \|\bm{\alpha_0}\|_2^2$ where  $\bm{\alpha}_0=[\alpha_{01},\alpha_{0i}]^T $. Table \ref{withpenalty} shows the estimated parameters.


\begin{table}[H]
\caption{Estimated parameters for regularized pairwise model}
\label{withpenalty}
\begin{center}
\begin{tabular}{ccccc}
\toprule 
pair& $\alpha_{01}$& $\alpha_{0i}$& $l_{01}$& $l_{0i}$\\
\midrule
($y_1,y_2$)&\textbf{ 8.27 e-7}& 1.91& 6.61& -1.21\\
\midrule
($y_1, y_3$)& \textbf{-3.27 e-6}& 0.93& 2.37& 1.77\\
\bottomrule 
\end{tabular}
\end{center}
\end{table}

\vspace{-1em}
One can directly observe from Table \ref{withpenalty} that when adding regularization, $\alpha_{01}$ is shrunk to nearly zero in both submodels. This implies that  $\operatorname{cov}_{1i}^f(d)\approx 0 \> \forall x$ and $i \in\{2,3\}$ as $K_{01}$ is almost identically zero and hence $y_1$ is predicted independently (for exact sparse solutions one can use SCAD or the $L^1$ norm). This not only confirms that regularization can limit information sharing but also illustrates that in the proposed $\mathcal{MGP}$ models, one can automatically perform variable selection (cluster the outputs that ought to be predicted independently). A user might then choose to perform a separate $\mathcal{MGP}$ on the selected subsets. To the best of our knowledge this is the first model to achieve simultaneous estimation and functional selection using dependent $\mathcal{GP}$s.  

\subsection{Illustration with Subsets of Correlated Outputs}
\subsubsection{Low Dimensional Setting} \label{lowdim}
For this setting we study the case when subsets of outputs are correlated. We first perform inference in a low dimensional regime to compare with the full $\mathcal{MGP}$ that does not face the challenge of large complexity and extremely high dimensional parameter space. We generate outputs $y^{(j)}_i(x) = f^{(j)}_i(x)+\epsilon^{(j)}_i(x)$ from
\begin{align*}
    &f^{(1)}_i(x) = x^2/(0.8\cdot(1-x)) \quad &i\in \{1,2\}\\
    &f^{(2)}_i(x) = x/(1-x) \quad &i\in \{3,4\}\\
    &f^{(3)}_i(x) = 2 \cdot x^2 \quad &i\in\{5,6\}\\
    &f^{(4)}_i(x) = x^3 \quad &i\in\{7,8\}
\end{align*}

\begin{figure}[!htbp]
	\centering
	\makebox[\linewidth]{\includegraphics[keepaspectratio=true,width=0.8\linewidth]{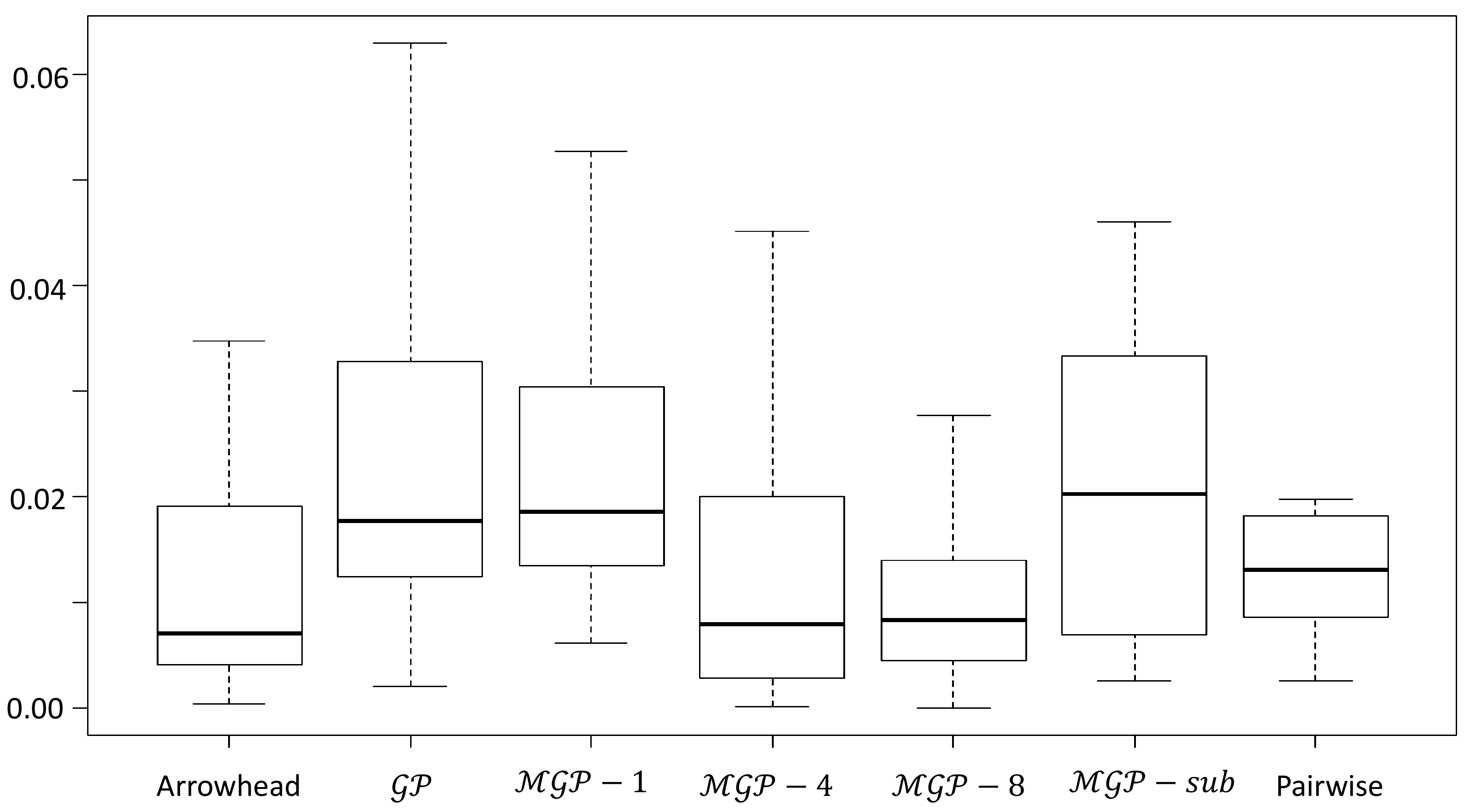}}
	\caption{Prediction comparison for $y_1(x)$ }
	\label{fig:lowdim}
\end{figure}

Here we focus on predicting outputs $y_1$, using the following models: (1) $\mathcal{MGP}-Q$ where $Q=1,4$ and $8$ respectively, (2) Pairwise model where predictions are combined using the robust product of experts in \citet{deisenroth2015distributed} (see Appendix F), (3) Arrowhead model, (4) a univariate $\mathcal{GP}$ on $y_1$, (5) A bivariate $\mathcal{GP}$ with outputs $y_1$ and $y_2$ (i.e. outputs in $y^{(1)}_i(x)$ ) denoted as $\mathcal{MGP}-sub$. We use $p=7$, $\sigma_i=0.1$ for $i=1,2$ and $\sigma_i=0.01$ for $i=3,4\cdots,8$. The squared exponential kernel is used. Results for the MSE over $p=30$ uniformly spaced points in $[0,0.8]$ are given in Fig. \ref{fig:lowdim} where the experiment is replicated 30 times. Also Tukey's multiple comparison test is done and only significant results are reported in the discussion below.  

The first result to observe is that $\mathcal{MGP}-sub$ outperformed $\mathcal{MGP}-1$ which confirms that negative transfer occurred since when outputs from $y^{(1)}_i$ are analyzed separately they produce better predictive results. However the key observation is that the pairwise and arrowhead models outperform $\mathcal{MGP}-1$. \textit{This is because, when learning an output from $y_i^{(1)}$, both pairwise and arrowhead models can leverage the correlation with other outputs and still avoid negative transfer evidenced through $\mathcal{MGP}-1$}. Also both proposed latent structures had comparable performance with $\mathcal{MGP}-8$, which confirms their capability to provide competitive predictive results with lower number of parameters and computational complexity. Another interesting result is that $\mathcal{MGP}-4$ and $\mathcal{MGP}-8$ have similar performance. Indeed, this is expected based on theorem \ref{thm2}, where if we have $M$ distinct subsets we only need $Q=M$ to avoid negative transfer. However in reality $M$ is not given in in advance.

\subsubsection{Moderate and Large Dimensional Setting}
\begin{figure}[!htbp]
	\centering
	\includegraphics[keepaspectratio=true,width=0.5\linewidth]{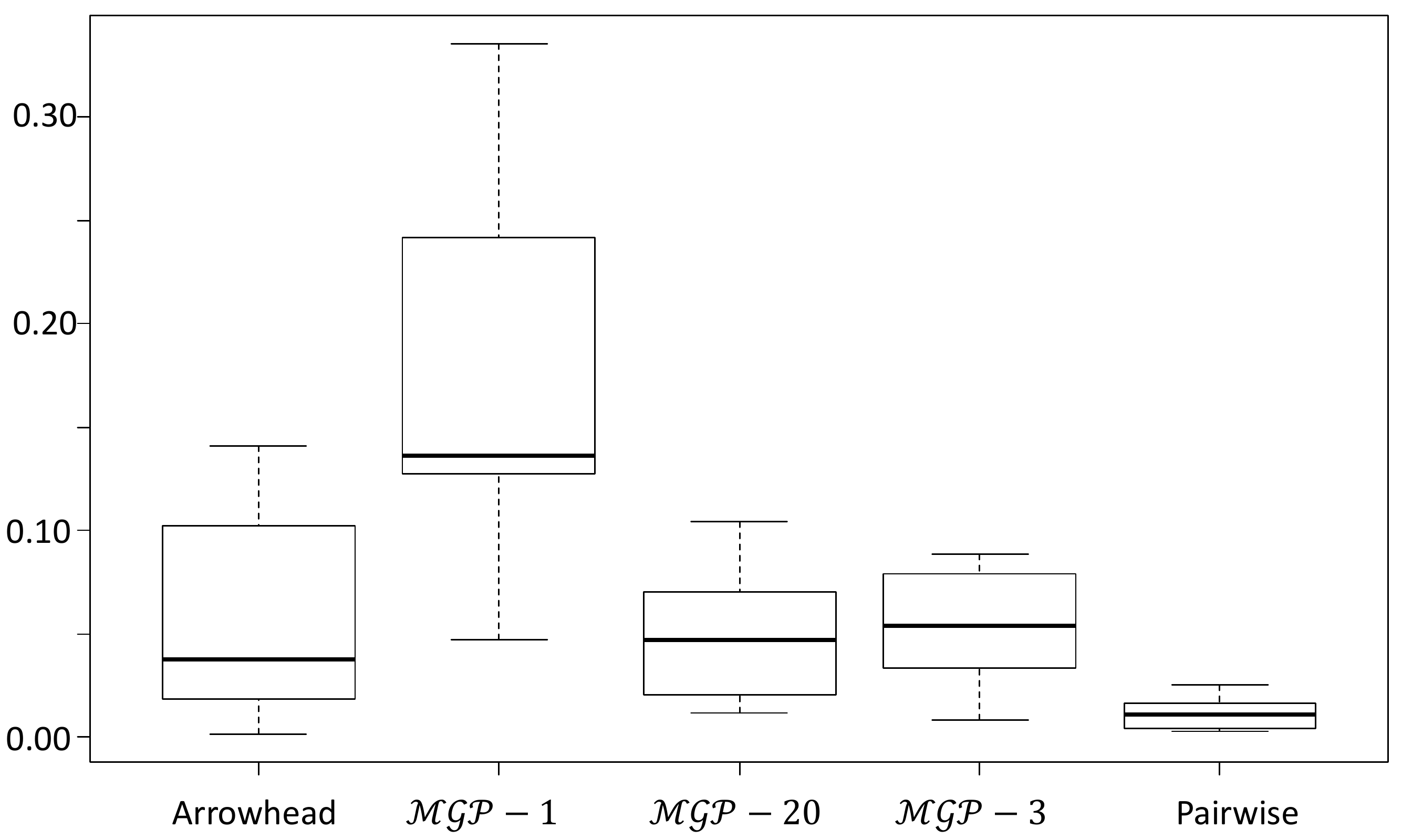}
	\caption{Moderately Large Parameter Space}
	\label{fig:moderate}
\end{figure}

\begin{figure}[!htbp]
	\centering
	\includegraphics[keepaspectratio=true,width=0.5\linewidth]{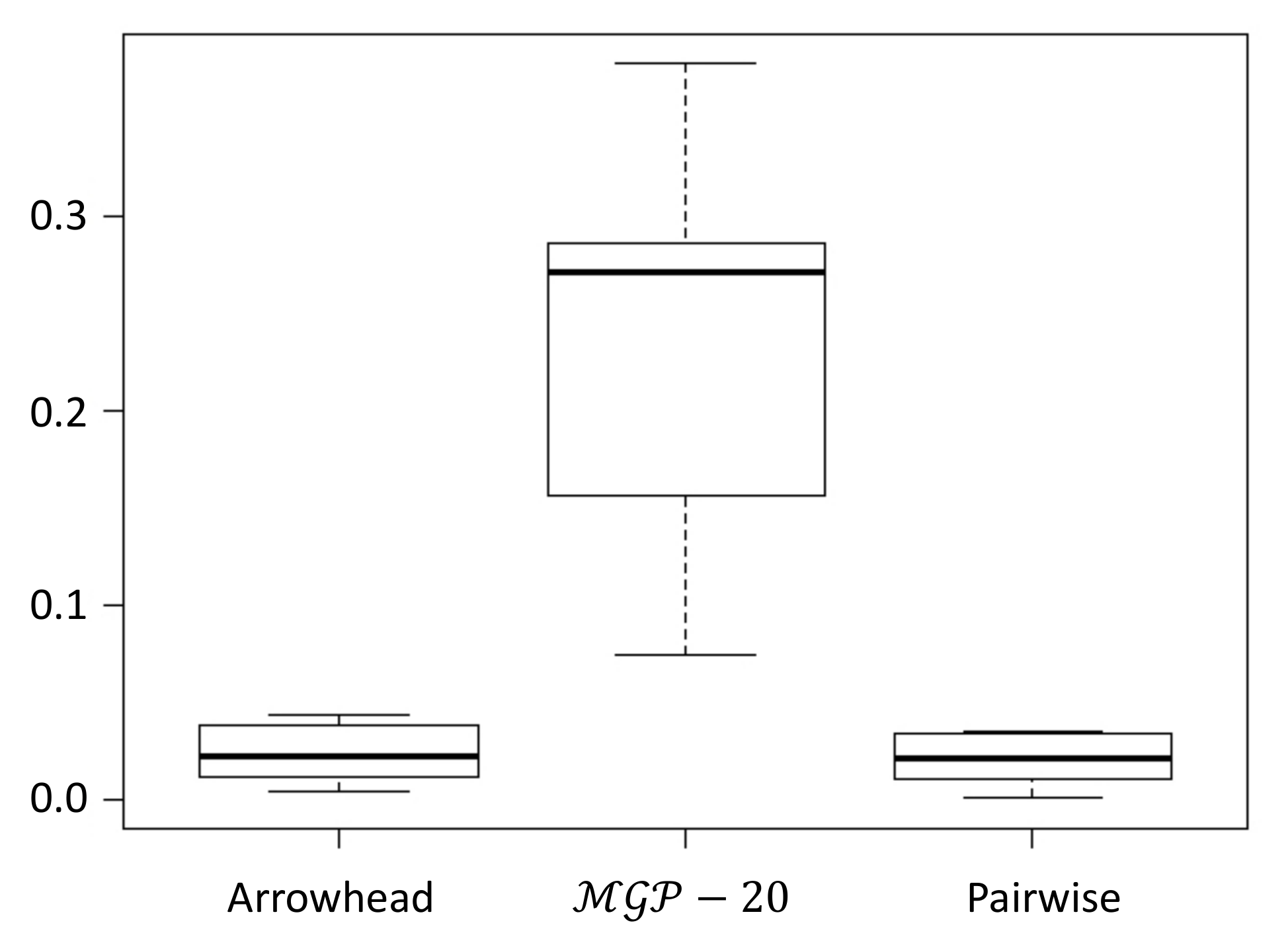}
	\caption{Large Parameter Space}
	\label{fig:large}
\end{figure}

In this setting we aim to compare our proposed structures when the number of parameters is significantly increased. Specifically $N=20$ and $N=50$ outputs are used. For the $N=20$ settings, outputs are generated from a $\mathcal{GP}$ with zero mean and $\mbox{cov}^y_{ii}(x,x^{\prime})= \alpha_i^2\exp{\dfrac{(x-x')^2}{2\cdot l_i^2}}+\sigma^2_i \delta(x,x^{\prime})$ under the following settings: (We slightly abuse notation with $\delta$ being an indicator here) 
\begin{align*}
    &\alpha_i=4, l_i=1, \sigma_i = 0.005\quad \text{for} \quad i=1, \cdots, 5 \\
    &\alpha_i=1, l_i=4, \sigma_i = 0.0001\quad \text{for} \quad i=6, \cdots, 12\\
    &\alpha_i=4, l_i=1, \sigma_i = 0.001\quad \text{for} \quad i=13, \cdots, 20
\end{align*}
For the $N=50$, the settings can be found in Appendix G. We generate $p=15$ points evenly spaced in $[0,3]$ for each output where 8 points are used for training 7 for testing. Similar to the setting in Sec. \ref{lowdim} we test on $y_1$ under 30 replications. The results for $N=20$ are shown in Fig. \ref{fig:moderate}. From the result we can see that $\mathcal{MGP}-3$, $\mathcal{MGP}-20$ and the arrowhead model yield similar results (also confirmed via Tukey's test). This once again confirms that the arrowhead model has competitive performance and that with enough latent functions one can avoid negative transfer. Yet the interesting result is the fact that the pairwise model showed much better performance. The reason is due to the fact that in each pair the number of estimated parameters is very small and thus one can except better estimators compared with competing models as parameter dimension increases. This fact is further illustrated through the results of $N=50$ shown in Fig. \ref{fig:large}.
In Fig. \ref{fig:large} for the $\mathcal{MGP}$ we use $Q=20$ thus we have $QN\omega+N=2050$ parameters to estimate. The results show that with $N=50$ there is a huge decrease in predictive performance. This result is expected as it is extremely challenging to obtain good parameter estimates specifically for a $\mathcal{GP}$ likelihood function which is known to be highly non-linear with many local critical point with bad generalization power. Indeed, similar decrease in performance in a high dimensional parameter space has been reported in  \citet{li2016pairwise} and \citet{li2018pairwise}. Here both the arrowhead and pairwise models offer a solution that, not only scales with any $N$, but also can lead to better performance. 

We note that on average $\mathcal{MGP-}20$ with $N=50$ took $\approx 24$ hours to estimate despite the computational complexity being relatively small with $P=400$. While the arrowhead model took $\approx 30$ minutes and $\approx 30$ seconds for the pairwise model. In practice its very common to have $N>>50$, this indeed exacerbates the challenge above and further highlights the needs for the proposed relaxation models.

\subsection{Case Studies}

\begin{figure}[!htbp]
	\centering
	\makebox[\linewidth]{\includegraphics[keepaspectratio=true,width=\linewidth]{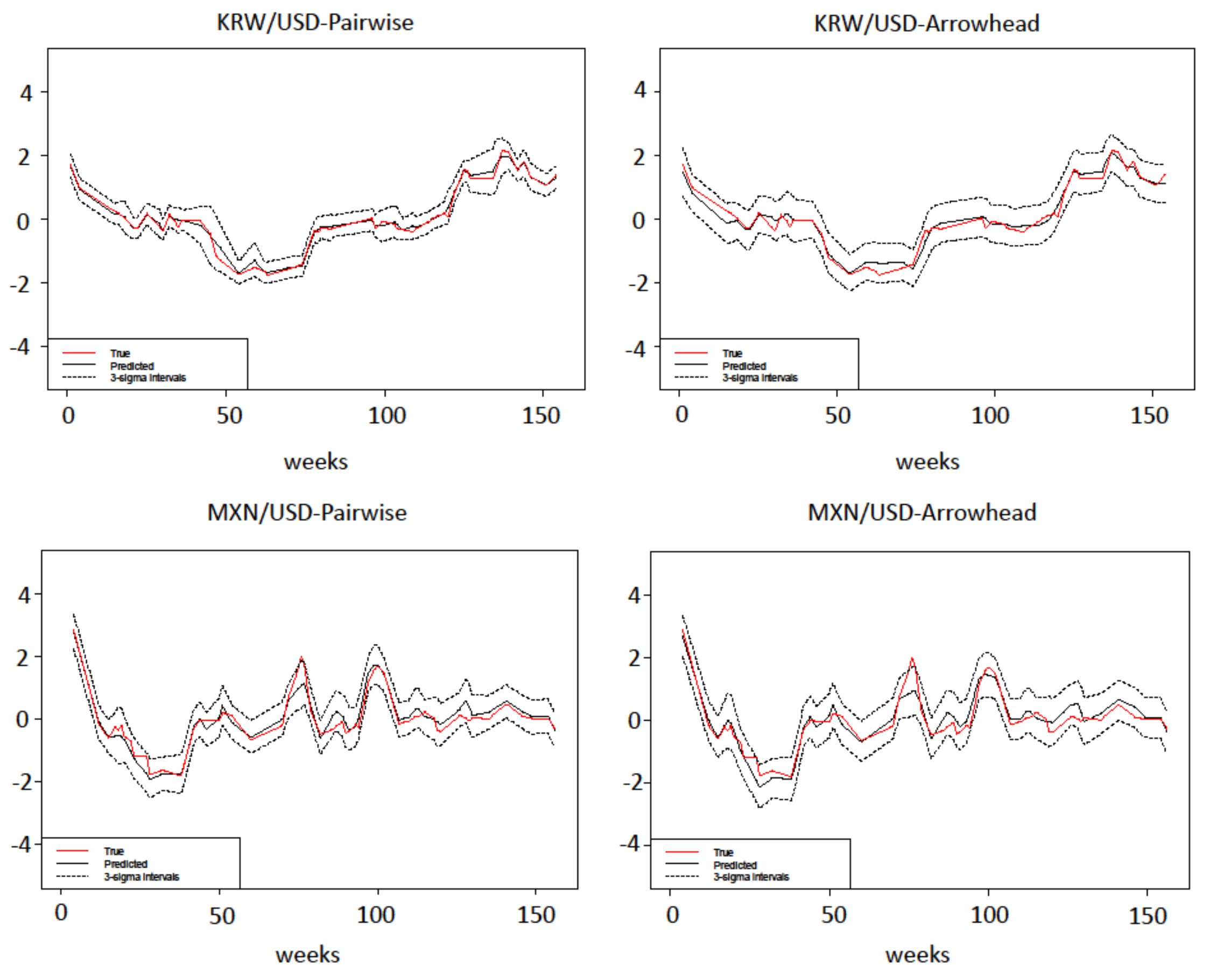}}
	\caption{illustartions on exchange rate data }
	\label{KRW}
\end{figure}

In this setting, we use the real data set from the pacific exchange rate service (\url{http://
	fx.sauder.ubc.ca/data.html}). Our goal is to predict the foreign exchange rate compared to the United States dollar currency. We utilized the exchange rates of the top ten international currencies 
during the 157 weeks of from 2017 to 2020.  Each output is adjusted to have zero mean and unit variance. We randomly choose 100 points as training points and the remaining 57 points as testing points. Table \ref{MXNtable}  shows the predictive arrow using different models while Fig. \ref{KRW} provides illustartions. Once again the results confirm the 
need for the proposed relaxation models as parameter estimates tend to deteriorate as the parameter dimension increases.  
\vspace{-1em}
\begin{table}[H]
\caption{Predictive Error of MXN/USD and KRW/USD}
\label{MXNtable}
\begin{center}
\begin{tabular}{cccc}
\toprule 
model& pairwise& arrowhead& $\mathcal{MGP-}10$\\
\midrule
MSE(MXN/USD)&0.040& 0.031& 0.217\\
\midrule
MSE(KRW/USD)& 0.015& 0.035& 0.322\\
\bottomrule 
\end{tabular}
\end{center}
\end{table}
\vspace{-1em}

Note that analysis on a Parkinson dataset to predict a disease symptom score is provided in Appendix G. The dataset is available on \url{http://archive.ics.uci.edu/ml/datasets/Parkinsons+Telemonitoring}.


\section{Conclusion} \label{sec:conclusion}
This article addresses the key challenge of constructing an $\mathcal{MGP}$ that can borrow strength across outputs without forcing correlation which can lead to negative transfer.  We show that this is achieved by having a sufficient number of latent functions regardless of the kernel used. This however comes with the challenge of greatly augmenting the parameter space. To this end, we propose two latent structures that can avoid negative transfer and maintain estimation in a low-dimensional parameter space. A key feature of our structures is that they allow functional variable selection via regularization. Further analysis into the use of such latent structures and other dependent $\mathcal{GP}$ models for selection in functional data settings or probabilistic graphical models can be an interesting topic to explore.

\nocite{langley00}

\bibliography{Negative_Latent_cite}
\bibliographystyle{icml2019}

\clearpage
\twocolumn[
\icmltitle{Supplementary materials:\\On Negative Transfer and Structure of Latent Functions\\ in Multi-output Gaussian Processes}]
\begin{appendices}
      \section{Proof of Lemma 1}
Consider the $\mathcal{MGP}$ model with 2 outputs $y_1$ and $y_2$, modeled with one latent function $X_1$, where $f_i(\bm{x}) = K_{1i}(\bm{x})\star X_1(\bm{x})$ and $X_1$ is Dirac Delta function. We will show that for any $\bm{x}, \bm{x'}\in \mathcal{R^D}$, $\mbox{cov}^f_{12}(\bm{x}, \bm{x}^{\prime}) $ = 0 if and only if at least one of $K_{11}$, $K_{12}$ is identically equal to zero. The sufficiency is obvious, then we prove necessity.\\
First assume $K_{11}$ and $K_{12}$ satisfy the first condition, i.e. $\exists\ \alpha_{11}, \alpha_{12}\in \mathcal{R}$ such that $K_{11} = \alpha_{11}{k_{11}}$ and $K_{12} = \alpha_{12}{k_{12}}$, where ${k_{11}}> 0$ and ${k_{12}}> 0$. Gaussian, Matern, rational quadratic, periodic and locally periodic kernels are typical examples for this case. For any two inputs points $\bm{x}$ and $\bm{x'}$, denote $\bm{d}= \bm{x}-\bm{x}'$. Then 
\begin{align*}
    \mbox{cov}^f_{12}(\bm{x}, \bm{x}^{\prime}) &= \int_{-\infty}^{+\infty}K_{11}(\bm{u})K_{12}(\bm{u}-\bm{d})d\bm{u}\\
    & = \alpha_{11}\alpha_{12}\int_{-\infty}^{+\infty}{k_{11}}(\bm{u}){k_{12}}(\bm{u}-\bm{d})d\bm{u}
\end{align*}
Since ${k_{11}}> 0$ and ${k_{21}}> 0$, $\int_{-\infty}^{+\infty}{k_{11}}(\bm{u}){k_{12}}(\bm{u}-\bm{d})d\bm{u} \neq 0$ for $\forall \bm{d}\in \mathcal{R}^D$. Thus, $\mbox{cov}^f_{12}(\bm{x}, \bm{x}^{\prime}) =0$ if and only if at least one of $\alpha_{1i}$, $i=1,2$ is equal to zero, i.e. at least one of $K_{1i}$, $i=1,2$ is identically equal to 0.

Then consider the second case when $k_{1i}$ has the form $\sum_u {a_u^2} \text{exp}\big(\bm{x}^T\bm{B}_u\bm{x}\big)\text{cos}(2\pi\bm{c}_u\bm{x}^T)$ with parameters $({a_u}, \bm{B}_u,\bm{c}_u)$. Here for simplicity, we only prove for one dimension case when $x,x'\in \mathcal{R}$ and the proof for general case is similar. We rewrite $k_{1i}$, $i=1,2$ as $k_{1i} = \sum_{q=1}^{Q_i}{a_{iq}^2 \exp\{-\sigma_{iq}^2 d^2\}\cos(\mu_{iq}d)}$. Since now $k_{1i}({d}) = k_{1i}(-{d})$, 
\begin{align*}
    \mbox{cov}^f_{12}({x}, {x}^{\prime})  &= \int_{-\infty}^{+\infty}K_{11}({u})K_{12}({u}-{d})du\\
    & = \int_{-\infty}^{+\infty}{K_{11}}({u}){K_{12}}({d}-{u})d{u}\\
    & = K_{11}\star K_{12}({d})\\
    & = \mathcal{F}^{-1}(\mathcal{F}(K_{11})\cdot \mathcal{F}(K_{12}))({d})
\end{align*}
where $\mathcal{F}$ is the Fourier operator and $\star$ denote the convolution operator. The last equality is derived using the conclusion of Convolution Theorem. Hence $\mbox{cov}^f_{12}({x}, {x}^{\prime}) =0$ if and only if $$\mathcal{F}(K_{11})\cdot \mathcal{F}(K_{12})(\xi) = \sum_{k=1}^{Q_1}\sum_{j=1}^{Q_2}\dfrac{a_{1k}^2a_{2j}^2}{4}M_{1k}M_{2j}= 0$$
where 
\begin{align*}
    M_{1k} &= \sqrt{\dfrac{\pi}{\sigma_{1k}^2}}(\exp(\dfrac{(\mu_{1k}-2\pi \xi)^2}{-4\sigma_{1k}^2})+\exp(\dfrac{(\mu_{1k}+2\pi \xi)^2}{-4\sigma_{1k}^2}))\\
    M_{2j} &= \sqrt{\dfrac{\pi}{\sigma_{2j}^2}}(\exp(\dfrac{(\mu_{2j}-2\pi \xi)^2}{-4\sigma_{2j}^2})+\exp(\dfrac{(\mu_{2j}+2\pi \xi)^2}{-4\sigma_{2j}^2}))
\end{align*}
Since $ M_{1k}, M_{2j}> 0$, we have $a_{1k}^2a_{2j}^2=0$ for any $k\in\{1,2,\cdots Q_1\}$ and  $j\in\{1,2,\cdots Q_2\}$. Therefore, we reach the conclusion either $a_{1k} =0,\forall{k}$ or $a_{2j} =0,\forall{j}$, i.e. at least one of $K_{1i}$, $i=1,2$ is identically equal to 0. \\
For general case when $X_1$ is a $\mathcal{GP}$ constructed from $\mathcal{CP}$, i.e. $$\operatorname{cov}(X_1(\bm{u}),X_1(\bm{u}')) = \int_{-\infty}^{+\infty}K_{X_1}(\bm{v})K_{X_1}(\bm{v}-\bm{d})d\bm{v}$$where $\bm{d}= \bm{u}-\bm{u}'$. Consider the first case when there $\exists\ \alpha_{11}, \alpha_{12}, \alpha_1 \in \mathcal{R}$ such that $K_{11} = \alpha_{11}{k_{11}}$, $K_{12} = \alpha_{12}{k_{12}}$ and $K_{X_1} = \alpha_{1}{k_{X_1}}$, where ${k_{11}}> 0$, ${k_{12}}> 0$ and ${k_{X_1}}> 0$.
\begin{align*}
    &\quad \operatorname{cov}_{12}^f(\bm{x},\bm{x}') \\ &=  \int_{-\infty}^{+\infty}K_{11}(\bm{x}-\bm{z})\int_{-\infty}^{+\infty}K_{12}(\bm{x}'-\bm{z}')\cdot \\ &\quad \int_{-\infty}^{+\infty}K_{X_1}(\bm{v})K_{X_1}(\bm{v}-\bm{d})d\bm{v}d\bm{z}'d\bm{z}\\ & = \alpha_{11}\alpha_{12}\alpha_1\int_{-\infty}^{+\infty}k_{11}(\bm{x}\!-\!\bm{z})\int_{-\infty}^{+\infty}k_{12}(\bm{x}'\!-\!\bm{z}')\cdot \\ &\quad \int_{-\infty}^{+\infty}k_{X_1}(\bm{v})k_{X_1}(\bm{v}\!-\!\bm{d})d\bm{v}d\bm{z}'d\bm{z}
\end{align*}
Similar as the argument when $X_1$ is Dirac Delta function, we have $\operatorname{cov}_{12}^f(\bm{x}, \bm{x}')=0$ if and only if one of $K_{11}, K_{12}$ and $K_{X_1}$ is identically equal to 0. \\
Now consider the case when $K_{11}$, $K_{12}$ and $K_{X_1}$ satisfy the second condition, then 
\begin{align*}
    & \quad \operatorname{cov}_{12}^f(\bm{x}, \bm{x}')\\ 
    & = K_{11}\star K_{12}\star K_{X_1}\star K_{X_1}(\bm{d})\\
    & = \mathcal{F}^{-1}(\mathcal{F}(K_{11}\star K_{12})\cdot \mathcal{F}(K_{X_1}\star K_{X_1}))(\bm{d})\\
    & = \mathcal{F}^{-1}(\mathcal{F}(K_{11})\cdot \mathcal{F}(K_{12})\cdot \mathcal{F}(K_{X_1})\cdot\mathcal{F}(K_{X_1}))(\bm{d})\\
\end{align*}
Hence $\operatorname{cov}_{12}^f(\bm{x}, \bm{x}')=0$ if and only if $\mathcal{F}(K_{11})\cdot \mathcal{F}(K_{12})\cdot \mathcal{F}(K_{X_1})\cdot\mathcal{F}(K_{X_1})=0$. Similar as the proof when $X_1$ is Dirac Delta function, we reach the conclusion that $\operatorname{cov}_{12}^f(\bm{x}, \bm{x}')=0$ if and only if one of $K_{11}, K_{12}$ and $K_{X_1}$ is identically equal to 0.

\section{Proof of Theorem 2}
We use an induction argument to establish the proof.\\In Lemma 1, we have shown that if we model two outputs using one latent function, then $\operatorname{cov}_{12}^f(\bm{x}, \bm{x}')\neq 0 $ for any $\bm{x}, \bm{x'}\in \mathcal{R}^d$. On the other hand, if we use $Q(Q\geq 2)$ latent functions, then the model has enough flexibility to construct $f_i$, $i=1,2$ from different latent functions, i.e. $f_1 = K_1\star X_1$ and $f_2 = K_2\star X_2$, where $X_1\neq X_2$. In this case, $\operatorname{cov}_{12}^f(\bm{x}, \bm{x}') =0$ for any $\bm{x}$ and $\bm{x'}$. 
 i.e. we have proved that when $M=2$, the model could achieve $$\operatorname{pr}\left(y_{i}\left(\boldsymbol{x}_{0}\right) | D_{\mathcal{I}}\right)=\operatorname{pr}\left(y_{i}\left(\boldsymbol{x}_{0}\right) | D_{\mathcal{I}_{m}}\right)\; \forall\ i\in \mathcal{I}_{m}, m=1,2$$ if and only if the number of latent function $Q\geq 2$.\\
Then we use induction: Assume the conclusion holds for $M= K-1$: Consider $\mathcal{MGP}$ with $N$ outputs $y_1, y_2\cdots, y_N$. Let the index set of all outputs $\mathcal{I}$ comprise of $K-1$ non-empty disjoint subsets $\mathcal{I}=\{\mathcal{I}_{1},\cdots, \mathcal{I}_{K-1}\}$. If for any $i,j\in \{1,2,\cdots, K-1\}$ and $\forall\ \bm{x}, \bm{x'} \in \mathcal{R}^D$, $\operatorname{Cov}(f_{i_s}(\bm{x}),f_{j_t}(\bm{x'}))=0$, then we at least need $K-1$ latent functions, where $i_s\in \mathcal{I}_i$ and $j_t\in \mathcal{I}_j$. Now consider $M=K$. We could separate this problem into two steps: first, we want $K-1$ disjoint subsets of $y_1,\cdots, y_N$ to be uncorrelated. Denote the index of these $K-1$ subsets as $\mathcal{I}_1, \mathcal{I}_2\cdots, \mathcal{I}_{K-1}$. Follow the assumption in the induction, we at least need $K-1$ latent functions $\{X_1,X_2, \cdots, X_{K-1}\}$, i.e. any output $f_{i_s}$ in $\mathcal{I}_i$, $i=1,2,\cdots, K-1$ is constructed from the convolution of $X_i$ and a smooth kernel: $f_{is} =  K_{i_s}\star X_i$, $i=1,2\cdots, K-1$. Then, we want the outputs with index $\mathcal{I}_K=\mathcal{I} \backslash \{\mathcal{I}_1\cup \mathcal{I}_2 \cdots \cup \mathcal{I}_{K-1}\}$ to be uncorrelated with the outputs in the previous $K-1$ subsets. If we still use $K-1$ latent functions, then the outputs $y_i$, $i\in \mathcal{I}_K$ has to be constructed using the latent functions in $\{X_1,X_2, \cdots, X_{K-1}\}$. Then, similar to the case when we have 2 outputs, there must exist a subset $\mathcal{I}_{i_0}$, $i_0\in \{1,2,\cdots, K-1\}$, such that $y_i$, $i\in \mathcal{I}_{i_0}$ has non-zero covariance function with the outputs in $\mathcal{I}_K$, i.e. these two subsets are correlated. On the other hand, if we use $K$ latent functions, then the model has capability to construct the outputs in $\mathcal{I}_i$, $i=1,2,\cdots, K$ from different latent functions, i.e. any output $f_{i_s}$, $i_s\in\mathcal{I}_i$ can be constructed as $f_{i_s} = K_{i_s}\star X_i$, $i=1,2,\cdots, K$. In this case,
$\operatorname{cov}(f_{i_s}(\bm{x}),f_{j_t}(\bm{x'})) =0$ for any $\bm{x}$ and $\bm{x'}$, where $f_{i_s}$ and $f_{j_t}$ are respectively arbitrary outputs with index in $\mathcal{I}_{i}$ and $\mathcal{I}_{j}$.\\ Therefore, we have proved that when $\mathcal{I} = \{\mathcal{I}_1\cup \mathcal{I}_2 \cdots \cup \mathcal{I}_{M}\}$, where $\mathcal{I}_1,\cdots, \mathcal{I}_M$ are $M$ non-empty disjoint subset of $\mathcal{I}$, $\operatorname{Cov}(f_{i}(\bm{x}),f_{j}(\bm{x'})) =0$ for any $\bm{x}$ and $\bm{x'}$ if and only if the number of latent functions $Q\geq M$, where $i\in \mathcal{I}_m$ and $j\in \mathcal{I}/\mathcal{I}_m$. That is to say, the model could achieve $$\operatorname{pr}\left(y_{i}\left(\boldsymbol{x}_{0}\right) | D_{\mathcal{I}}\right)=\operatorname{pr}\left(y_{i}\left(\boldsymbol{x}_{0}\right) | D_{\mathcal{I}_{m}}\right) \quad \forall i \in \mathcal{I}_{m}$$ if and only if $Q\geq M$. 

\section{Illustrative example of Theorem 2}
Here we present a simple example when $N=2$ and $Q=2$ to illustrate Theorem 2. We have proved that in this case, the model could achieve $$\operatorname{cov}_{12}^f(\bm{x}, \bm{x'}) = 0,\quad \forall \bm{x}, \bm{x'}\in \mathcal{R}^D$$
For any new input point $\bm{x^{\star}}$, the integrative analysis of $\bm{y}_1$ and $\bm{y}_2$ leads to the prediction:
\begin{align*}
\left(\begin{array}{cc} \bm{y}_1^{\star} \\ \bm{y}_2^{\star}\end{array}\right)&= \left(\begin{array}{cc} C_{11}^{\star}& C_{12}^{\star}\\ C_{21}^{\star} & C_{22}^{\star}\end{array}\right)\left(\begin{array}{cc}
C_{11} & C_{12} \\C_{21} & C_{22}
\end{array} \right)^{-1}\left(\begin{array}{cc}\bm{y}_1 \\\bm{y}_2 \end{array}\right)\\& = \left(\begin{array}{cc} C_{11}^{\star}& 0\\ 0 & C_{22}^{\star}\end{array}\right)\left(\begin{array}{cc}
C_{11} & 0 \\0 & C_{22}
\end{array} \right)^{-1}\left(\begin{array}{cc}\bm{y}_1 \\\bm{y}_2 \end{array}\right)\\& = \left(\begin{array}{cc} C_{11}^{\star}C_{11}^{-1}\bm{y}_1 \\C_{22}^{\star}C_{22}^{-1}\bm{y}_2\end{array}\right)
\end{align*} 
      where $\bm{y}_i = f_i(\bm{x})+\epsilon_i(\bm{x})$ and $\bm{y}_i^{\star} = f_i(\bm{x}^{\star})+\epsilon_i(\bm{x})$, $i=1,2$; $C_{ij} = \operatorname{cov}_{12}^f(\bm{x},\bm{x'})$, $C_{ij}^{\star} = \operatorname{cov}_{12}^f(\bm{x}^{\star},\bm{x'})$ 
      The prediction result is exactly the same with that when we model $y_1$ and $y_2$ independently, which means that the model has capability to make the $\mathcal{MGP}$ model collapse into two 2 independent $\mathcal{GP}s$, hence we achieve our goal of avoiding negative transfer between $\bm{y}_1$ and $\bm{y}_2$.

\section{Proof of Theorem 3}

To establish the proof we start with the fact that the unpenalized likelihood $\ell(\bm{\theta};\bm{y})$ is consistent such that $||\hat{\bm{\theta}}-\bm{\theta}^{\ast}||=O(r^{-1}_P)$ \cite{basawa1980statistical,basawa1976asymptotic}, under the usual regularity conditions similar to those of independent normal data (refer to chapter 7 of \citet{basawa1980statistical} and \citet{lehmann2006theory}). Our aim is to study $\ell_{\mathbb{P}}(\bm{\theta};\bm{y},\lambda)=\ell(\bm{\theta};\bm{y})+\mathbb{P}_\lambda(\bm{\theta}_0)$. 

The results here provide similar results to that of \citet{fan2001variable} yet under the correlated setting.  For consistent notation we maximize the log-likelihood with form $\ell_{\mathbb{P}+}(\bm{\theta})= \ell_+(\bm{\theta})-P\mathbb{P}_\lambda(\bm{\theta})$ where $\ell_+(\bm{\theta})=-\ell(\bm{\theta};\bm{y})=\mbox{log} \ \mathcal{L}(\bm{\theta};\bm{y})$. To prove theorem 2, we need to show that given $\varepsilon>0$ there exists a a constant $\mathcal{G}$ such that
\begin{align*}
pr \big( \ \underset{||\bm{g}||=\mathcal{G}}{\mbox{sup}} \ \ell_{\mathbb{P}+}(\bm{\theta}^{\ast}+r^{\ast} \bm{g}) < \ell_{\mathbb{P}+}(\bm{\theta}^{\ast}) \ \big)\geq 1-\varepsilon ,
\end{align*}
where $r^{\ast}=r^{-1}_P+r$. This implies that there exists a $\hat{\bm{\theta}}$ such that $||\hat{\bm{\theta}}-\bm{\theta}^{\ast}||=O(r^{\ast})=O(r^{-1}_P+r)$. We now have that
\begin{align*}
&\ell_{\mathbb{P}+}(\bm{\theta}^{\ast}+r^{\ast} \bm{g}) - \ell_{\mathbb{P}+}(\bm{\theta}^{\ast})=\big[ \ell_+(\bm{\theta}^{\ast}+r^{\ast} \bm{g})- \ell_+(\bm{\theta}^{\ast}) \big] \\
&  - P \sum_i \big[ \mathbb{P}_\lambda(|\theta^*_i+r^{\ast} g_{\theta_i}|) - \mathbb{P}_\lambda(|\theta^*_i|) \big]  ,
\end{align*}
\noindent
where $g_{\theta_i}$ denotes the element in $\bm{g}$ corresponding to $\xi_0$. Here we recall that the penalty is non-negative; $\mathbb{P}_\lambda(|\theta_i|)\geq 0 $, $\mathbb{P}_\lambda(0)= 0$ and $\mathbb{P}_\lambda(|\theta_i^{\prime}|) \geq \mathbb{P}_\lambda(|\theta_i|)$ if $|\theta_i^{\prime}| \geq |\theta_i|$. Therefore, $\ell_{\mathbb{P}+}(\bm{\theta}^{\ast}+r^{\ast} \bm{g}) - \ell_{\mathbb{P}+}(\bm{\theta}^{\ast}) \leq \big[ \ell_+(\bm{\theta}^{\ast}+r^{\ast} \bm{g})- \ell_+(\bm{\theta}^{\ast}) \big]$ when $\bm{\theta}^*=\bm{0}$. 

However, we have that $\ell_+(\bm{\theta}^{\ast}+r^{\ast} \bm{g})=\ell_+(\bm{\theta}^{\ast}) + r^* \ell_+^{\prime}(\bm{\theta}^{\ast})^{\top} \bm{g}-\frac{r^{2*}}{2}\bm{g}^{\top} \mathbb{I}(\bm{\theta}^{\ast})\bm{g}\{1+o(1)\}$ where $\ell_+^{\prime}(\bm{\theta}^{\ast})^{\top}$ is the gradient vector and $\mathbb{I}$ is the hessian at $\bm{\theta}^{\ast}$. Also,  $\mathbb{P}_\lambda(|\theta^*_i+r^{\ast} g_{\theta_i}|) - \mathbb{P}_\lambda(|\theta^*_i|)=r^* \mathbb{P}^{\prime}_\lambda(|\theta_i^*|) \mbox{sign}(\theta_i^*)g_{\theta_i^*} + r^{2*} \mathbb{P}^{\prime\prime}_\lambda(|\theta_i^*|)g_{\theta_i^*}^2\{1+o(1)\} \big]$. Therefore, 
\begin{align*}
\ell_{\mathbb{P}+}(\bm{\theta}^{\ast}+r^{\ast} \bm{g})& - \ell_{\mathbb{P}+}(\bm{\theta}^{\ast}) \\
& \leq r^* \ell_+^{\prime}(\bm{\theta}^{\ast})^{\top} \bm{g}-\frac{r^{2*}}{2}\bm{g}^{\top} \mathbb{I}(\bm{\theta}^{\ast})\bm{g}\{1+o(1)\}
\\&- P \sum_i \big[ r^* \mathbb{P}^{\prime}_\lambda(|\theta_i^*|) \mbox{sign}(\theta_i^*)g_{\theta_i^*}  \\
&+ r^{2*} \mathbb{P}^{\prime\prime}_\lambda(|\theta_i^*|)g_{\theta_i^*}^2\{1+o(1)\} \big] .
\end{align*}
The remainder of the proof is identical to theorem 1 in \citet{fan2001variable}.

\section{Formula of covariance functions using spectral kernels}
Consider the $\mathcal{MGP}$ model with  two outputs and one latent function $X_1$ $$f_{i}({x})= K_{1 i}({x}) \star X_{1}({x}),\quad i=1,2,\quad x\in \mathcal{R}$$  where 
\begin{align*}
    K_{1i}({d}) &= \sum_{q=1}^{Q_i}a_{qi}\cdot \exp\{-\sigma_{qi}^2{d}^2\}\cdot \cos(\mu_{qi} {d})
\end{align*}
From the proof of Lemma 1, we know that 
\begin{align*}
\operatorname{cov}_{12}^{f}\left({x}, {x}^{\prime}\right) &=\int_{-\infty}^{+\infty} K_{11}({u}) K_{12}({u}-{d}) d u \\
&=\int_{-\infty}^{+\infty} K_{11}({u}) K_{12}({d}-{u}) d {u} \\
&=K_{11} \star K_{12}({d}) \\
&=\mathcal{F}^{-1}\left(\mathcal{F}\left(K_{11}\right) \cdot \mathcal{F}\left(K_{12}\right)\right)({d})
\end{align*}K
Compute the Fourier Transform of $K_{11}$ and $K_{12}$ respectively: 
\begin{align*}
\mathcal{F}(K_{11})(\xi) & = \sum_{q=1}^{Q_1}\dfrac{a_{q1}}{2}\sqrt{\dfrac{\pi}{\sigma_{q1}^2}}(\exp\{-\dfrac{(\mu_{q1}-2\pi\xi)^2}{4\sigma_{q1}^2}\}+\\ & \quad \exp\{-\dfrac{(\mu_{q1}+2\pi\xi)^2}{4\sigma_{q1}^2}\})\\
& = \sum_{q=1}^{Q_1}\dfrac{a_{q1}}{2}M_{q1}(\xi)
\end{align*}
\begin{align*}
    \mathcal{F}(K_{12})(\xi) &= \sum_{q=1}^{Q_2}\dfrac{a_{q2}}{2}\sqrt{\dfrac{\pi}{\sigma_{q2}^2}}(\exp\{-\dfrac{(\mu_{q2}-2\pi\xi)^2}{4\sigma_{q2}^2}\}+\\ & \quad  \exp\{-\dfrac{(\mu_{q2}+2\pi\xi)^2}{4\sigma_{q2}^2}\})\\
    & = \sum_{q=1}^{Q_2}\dfrac{a_{q2}}{2}M_{q2}(\xi)
\end{align*}
Thus $$\mathcal{F}(K_1)(\xi)\cdot \mathcal{F}(K_2)(\xi) = \sum_{s=1}^{Q_1}\sum_{t=1}^{Q_2}\dfrac{a_{s1}a_{t2}}{4}M_{s1}M_{t2}$$
We hence get the covariance function between $f_1$ and $f_2$:
\begin{align*}
    \operatorname{cov}_{12}^f({x}, {x}') 
    & = \mathcal{F}^{-1}\left(\mathcal{F}\left(K_{11}\right) \cdot \mathcal{F}\left(K_{12}\right)\right)({d})\\
    & = \sum_{s=1}^{Q_1}\sum_{t=1}^{Q_2} \dfrac{a_{s1}a_{t2}}{2}\sqrt{\dfrac{\pi}{\sigma_{s1}^2+\sigma_{t2}^2}}H({d})
\end{align*}
where 
\begin{align*}
    H({d}) &= (e^{A_1({d})}\cos{(\theta_1{d})}+e^{A_2({d})}\cos{(\theta_2{d})})\\
    A_1({d}) &= \dfrac{-(\mu_{s1}-\mu_{t2})^2-4\sigma_{s1}^2\sigma_{t2}^2\pi^2{d}^2}{4(\sigma_{s1}^2+\sigma_{t2}^2)} \\
    A_2({d}) &= \dfrac{-(\mu_{s1}+\mu_{t2})^2-4\sigma_{s1}^2\sigma_{t2}^2\pi^2{d}^2}{4(\sigma_{s1}^2+\sigma_{t2}^2)} \\
    \theta_1 &= \dfrac{\mu_{s1}\sigma_{t2}^2+\mu_{t2}\sigma_{s1}^2}{\sigma_{s1}^2+\sigma_{t2}^2} \\
    \theta_2 &= \dfrac{\mu_{s1}\sigma_{t2}^2-\mu_{t2}\sigma_{s1}^2}{\sigma_{s1}^2+\sigma_{t2}^2}
\end{align*}
Note that in our simulation, we use the kernel with $Q_1=Q_2=1$ and the number of latent functions is $Q$. Thus the covariance function between $f_i$ and $f_j$ becomes 
$$
    \operatorname{cov}_{ij}^f({x}, {x}')  = \sum_{k=1}^{Q} \dfrac{a_{qi}a_{qj}}{2}\sqrt{\dfrac{\pi}{\sigma_{qi}^2+\sigma_{qj}^2}}H({d})
$$
\section{Case Study}
\subsection{$N=50$ Setting}
For $N=50$ setting, we generate 50 outputs from the Gaussian Process with mean zero and $\operatorname{cov}_{i i}^{y}\left(x, x^{\prime}\right)=\alpha_{i}^{2} \exp \frac{\left(x-x^{\prime}\right)^{2}}{2 \cdot l_{i}^{2}}+\sigma_{i}^{2} \delta\left(x, x^{\prime}\right)$ under the following setting
\begin{align*}
&\alpha_i=4, l_i=1, \sigma_i = 0.001\quad \text{for}\quad i=1,2,\cdots,9\\
&\alpha_i=1, l_i=4, \sigma_i = 0.0001\quad \text{for}\quad i=10,\cdots,19\\
&\alpha_i=1, l_i=8, \sigma_i = 0.0001\quad \text{for}\quad i=20,\cdots ,29\\
&\alpha_i=8, l_i=1, \sigma_i = 0.001\quad \text{for}\quad i=30,\cdots ,39\\
&\alpha_i=3, l_i=1, \sigma_i = 0.005\quad \text{for}\quad i=40,\cdots ,50
\end{align*}
We have 15 points evenly spaced in $[0,3]$ and we randomly choose 8 points from them as training data and the left 7 points are testing points. For the full MGP model in this setting, we use 20 latent functions to construct the model.

\subsection{Parkinson Data}
\begin{figure}[!htbp]
	\centering
	\makebox[\linewidth]{\includegraphics[keepaspectratio=true,width=\linewidth]{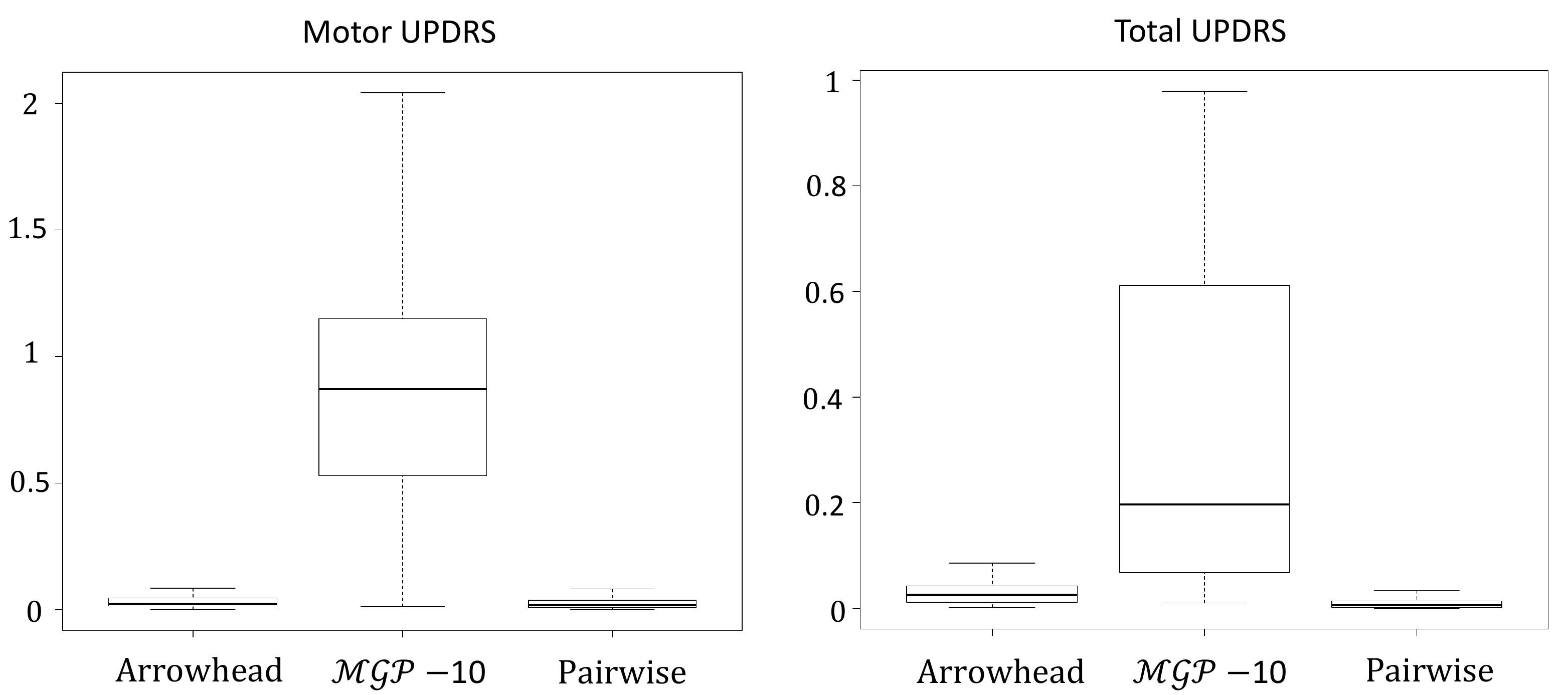}}
	\caption{Parkinson data set}
	\label{parkinson}
\end{figure}

We use the Parkinson data set to predict the disease symptom score(motor UPDRS and total UPDRS) of Parkinson patients at different times. The data set is available on \url{http://archive.ics.uci.edu/ml/datasets/Parkinsons+Telemonitoring}. At each time, we randomly choose 10 patients from the data set to model a $\mathcal{MGP}$ model with 10 outputs and randomly split 60\% data of each patient as training sets and 40\% as testing sets. Our goal is to predict the motor UPDRS and total UPDRS of the 10th patient in each round. We run our model for 70 times. Figure \ref{parkinson} shows the predictive error using different models. 

\end{appendices}

\end{document}